\documentclass[runningheads]{llncs}
\usepackage[T1]{fontenc}
\usepackage{graphicx}
\usepackage{xcolor}
\usepackage[export]{adjustbox}
\usepackage{multirow}
\usepackage{tabularx}
\usepackage{pifont}
\usepackage{amsmath}
\usepackage{bm}
\usepackage{geometry}
\usepackage{subfigure}
\usepackage{booktabs}
\usepackage{textcomp}
\usepackage{stfloats}
\usepackage{url}
\usepackage{verbatim}
\usepackage{mathtools}
\newcommand{\xmark}{\text{\ding{55}}}
\newcommand{\cmark}{\text{\ding{51}}}
\def\eg{\emph{e.g.}}
\usepackage{graphicx}   

\newcommand{\orcidlink}[1]{%
    \href{https://orcid.org/#1}{\includegraphics[width=10pt]{orcid_icon.png}}%
}

\begin{document}
\title{FocDepthFormer: Transformer with latent LSTM for \\ Depth Estimation from Focal Stack}
%
\author{Xueyang Kang\inst{1,2}\thanks{Corresponding author: \email{alexander.kang@tum.de}} \and
Fengze Han\inst{3} \and 
Abdur R. Fayjie\inst{1} \and 
Patrick Vandewalle\inst{1} \and 
Kourosh Khoshelham\inst{2} \and 
Dong Gong\inst{4}}
\authorrunning{X. Kang et al.}
%
\institute{Department of ESAT, KU Leuven, Leuven Belgium. \and Faculty of Engineering IT, the University of Melbourne, Melbourne, Australia. \and EI, Faculty, Technical University of Munich, Munich, Germany. \and EI Faculty, the University of New South Wales, Kensington, Australia.}
\maketitle              
\begin{abstract}
Most existing methods for depth estimation from a focal stack of images employ convolutional neural networks (CNNs) using 2D or 3D convolutions over a fixed set of images. However, their effectiveness is constrained by the local properties of CNN kernels, which restricts them to process only focal stacks of fixed number of images during both training and inference. This limitation hampers their ability to generalize to stacks of arbitrary lengths. To overcome these limitations, we present a novel Transformer-based network, FocDepthFormer, which integrates a Transformer with an LSTM module and a CNN decoder. The Transformer's self-attention mechanism allows for the learning of more informative spatial features by implicitly performing non-local cross-referencing. The LSTM module is designed to integrate representations across image stacks of varying lengths. Additionally, we employ multi-scale convolutional kernels in an early-stage encoder to capture low-level features at different degrees of focus/defocus. By incorporating the LSTM, FocDepthFormer can be pre-trained on large-scale monocular RGB depth estimation datasets, improving visual pattern learning and reducing reliance on difficult-to-obtain focal stack data. Extensive experiments on diverse focal stack benchmark datasets demonstrate that our model outperforms state-of-the-art approaches across multiple evaluation metrics.
\end{abstract}
\section{Introduction}
With the advancement of deep neural networks (DNNs) and the availability of high-volume data, the challenging task of depth estimation from monocular images \cite{guo2018learning} has seen significant success on benchmark datasets \cite{geiger2012we}.  However, the focal stack depth estimation problem, which is distinct from monocular depth estimation \cite{monodepth2}, \cite{meng2021cornet}, \cite{hornauer2022gradient}, stereo depth or disparity estimation \cite{godard2017unsupervised}, and multi-frame depth estimation \cite{schonberger2016structure}, yet has not received as much attention in the research community.

DDepth estimation using focus and defocus techniques involves predicting the depth map from a captured \emph{focal stack} of the scene, which consists of images taken at different focal planes \cite{xiong1993depth}. This problem, also known as depth of field control \cite{pentland1987new}, typically uses images obtained with a light field camera \cite{liu2017light}. Traditional methods \cite{suwajanakorn2015depth} rely on handcrafted sharpness features, but they frequently struggle in textureless scenes. To enhance feature extraction, Convolutional Neural Networks (CNNs) have been used to predict depth maps from focal stacks \cite{hazirbas2018deep}. Models like DDFFNet \cite{hazirbas2018deep}, AiFDepthNet \cite{wang2021bridging}, and DFVNet \cite{yang2022deep} leverage in-focus cues, while DefocusNet \cite{maximov2020focus} learns permutation invariant defocus cues, or Circle-of-Confusion (CoC). While these methods use 2D or 3D convolutions to represent visual and focal features, they are limited to processing focal stacks with a fixed number of images, which restricts their generalization to stacks of arbitrary length.
\begin{figure*}[tbph!]
\vspace{-0.6em}
\centering
\includegraphics[trim=0.0cm 1.6cm 1.5cm 0.86cm, width=0.90\textwidth]{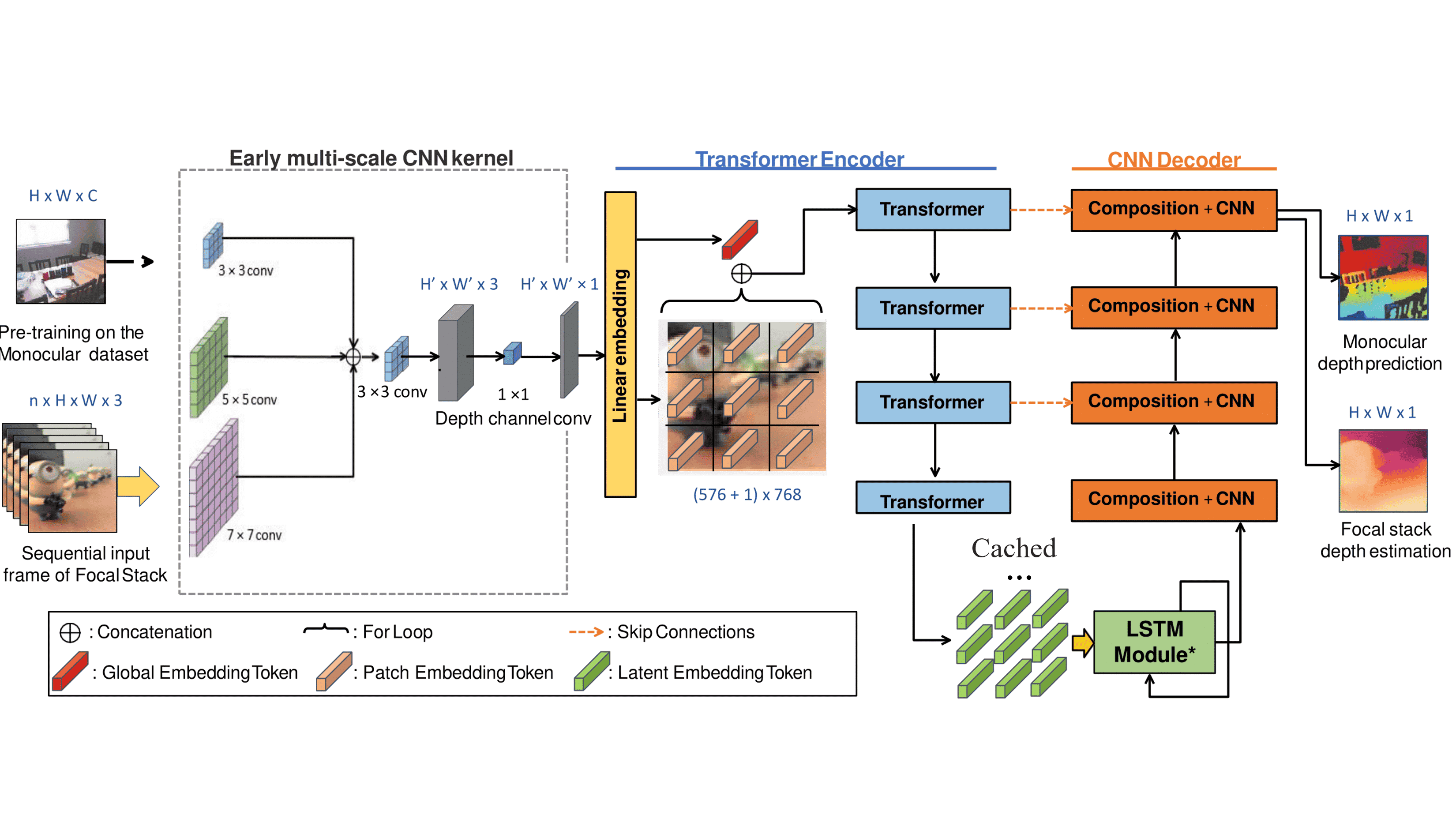}
\vspace{-1.0em}
\caption{The overview of our proposed network, FocDepthFormer, is presented with its core components: the Transformer encoder, the recurrent LSTM module, and the CNN decoder. Preceding the Transformer encoder, early-stage multi-scale convolutional kernels are depicted within the dashed line. The resulting multi-scale feature maps are concatenated and subjected to spatial and depth-wise convolution. Subsequently, the fused feature map of a image stack is divided into patches, which are then individually projected by a linear embedding layer into tokens. A red token represents a global embedding token mapped from the entire image and is summed with each individual patch embedding token.} 
\vspace{-0.4em}
\label{fig:model_arch}  
\end{figure*}

In this paper, we introduce FocDepthFormer, a novel network for depth estimation from focal stacks that combines LSTM and Transformer architectures. The core component is a module consisting of a Transformer encoder \cite{dosovitskiy2020image}, an LSTM-based recurrent module \cite{hochreiter1997long} applied to latent tokens, and a CNN decoder. The Transformer and LSTM models process spatial and stack information separately. Unlike CNNs, which are restricted to local representation, the Transformer encoder captures visual features with a larger receptive field. Given that focal stacks may have arbitrary and unknown numbers of images, we use the LSTM in the latent feature space to fuse focusing information across the stack for depth prediction. This approach differs from existing focal stack depth estimation methods \cite{yang2022deep,wang2021bridging,maximov2020focus} and monocular depth estimation methods based on CNNs or Transformers \cite{agarwal2022depthformer}, which typically handle inputs with a fixed number of images. Specifically, we compactly fuse activated token features via the recurrent LSTM module after the Transformer encoder,  enabling the model to handle focal stacks of arbitrary lengths during both training and testing, thereby providing greater flexibility. Before inputting data into the Transformer, we employ an early-stage convolutional encoder with multi-scale kernels \cite{xiao2021early} to capture low-level focus/defocus features across different scales. Considering the limited availability of focal stack data, our model enhances its representation of scene features through pre-training on monocular depth estimation datasets. Meanwhile, the recurrent LSTM module facilitates the model's ability to accommodate varying numbers of input images in focal stack.

The main contributions of this work are as follows:
\begin{itemize}
    \item We introduce a novel Transformer-based network model for depth estimation from focal stack images. The model uses a Vision Transformer encoder with self-attention to capture non-local spatial visual features, effectively representing sharpness and blur patterns. To accommodate an arbitrary number of input images, we incorporate a recurrent LSTM module. This structural flexibility allows for pre-training with monocular depth estimation datasets, thereby reducing the reliance on focal stack data.
    \item To fuse the stack features, we utilize the LSTM and implement a grouping operation to manage recurrent complexity across tokens, avoiding an increase in complexity as the token count grows with larger stack sizes. This is accomplished by applying the LSTM exclusively to a subset of activated embedding tokens while maintaining the information on other non-activated tokens through averaging aggregation. 
    \item We propose the use of multi-scale kernels in an early-stage convolutional encoder to enhance the capture of low-level focus/defocus cues at various scales.
\end{itemize}
\section{Related work}
\label{sec:rel}
\noindent\textbf{Depth from Focus/Defocus.} Depth estimation from focal stacks relies on discerning relative sharpness within the stack of images for predicting depth. Traditional machine learning methods \cite{xiong1993depth} treat this problem as an image filtering and stitching process. Johannsen \textit{et al.} \cite{johannsen2017taxonomy} provide a comprehensive overview of methods addressing the challenges posed by light field cameras, laying a foundation for research in this direction. More recently, CNN-based approaches have emerged in the context of focal stacks. DDFFNet \cite{hazirbas2018deep} introduces the first end-to-end learning model trained on the DDFF 12-Scene dataset. DFVNet \cite{yang2022deep} utilizes the first-order derivative of volume features within the stack. AiFNet \cite{wang2021bridging} aims to bridge the gap between supervised and unsupervised methods, accommodating both ground truth depth and its absence. Barratt \textit{et al.} \cite{shane20allinfocus} formulate the problem as an inverse optimization task, utilizing gradient descent search to simultaneously recover an all-in-focus image and depth map. DefocusNet \cite{maximov2020focus} exploits the Circle-of-Confusion, a defocus cue determined by focal plane depth, for generating intermediate defocus maps in the final depth estimation. Anwar \textit{et al.} \cite{anwar2021deblur} leverage defocus cues to recover all-in-focus images by eliminating blur in a single image. Recently, the DEReD model \cite{si2023fully} learns to estimate both depth and all-in-focus (AIF) images from focal stack images in a self-supervised way by incorporating the optical model to reconstruct defocus effects. Gur and Wolf \cite{gur2019single} present depth estimation from a single image by leveraging defocus cues to infer disparity from varying viewpoints.

\noindent\textbf{Attention-Based Models.} The success of attention-based models \cite{vaswani2017attention} in sequential tasks has led to the rise of the Vision Transformer for computer vision tasks. The Vision Transformer represents input images as a series of patches ($16 \times 16$). While this model performs well in image recognition compared to CNN-based models, a recent study \cite{xiao2021early} demonstrates that injecting a small convolutional inductive bias in early kernels significantly enhances the performance and stability of the Transformer encoder. In the context of depth estimation, Ranftl \textit{et al.} \cite{ranftl2021vision} utilize a Transformer-based model as the backbone to generate tokens from images, assembling these tokens into an image-like representation at multiple scales. DepthFormer \cite{agarwal2022depthformer} merges tokens at different layer levels to improve depth estimation performance. The latest advancement in this domain, the Swin Transformer \cite{liu2021swin}, achieves a larger receptive field by shifting the attention window, revealing the promising potential of the Transformer model.

\noindent\textbf{Recurrent Networks.} Recurrent networks, specifically LSTM \cite{hochreiter1997long}, have found success in modeling temporal distributions for video tasks such as tracking \cite{nwoye2019weakly} and segmentation \cite{xu2018youtube}. The use of LSTM introduces minimal computational overhead, as demonstrated in SliceNet \cite{pintore2021slicenet}, where multi-scale features are fused for depth estimation from panoramic images. Some recent works \cite{hutchins2022block} combine LSTM with Transformer for language understanding via long-range temporal attention.

\section{Method}

Given a focal stack $\mathbf{S}$, containing $N$ images ordered from near to far by focus distance, denoted as $\mathbf{S} = (\mathbf{x}_{i})_{i=1}^N$, where $\mathbf{x}_{i} \in \mathbf{R}^{H\times W \times 3}$ represents each single image, our objective is to generate a single depth map $\mathbf{D}\in\mathbf{R}^{H\times W \times 1}$ for a stack of images. In contrast to the vanilla Transformer \cite{dosovitskiy2020image}, we initially encode each image $\mathbf{x}$ using an \emph{early-stage multi-scale kernel-based convolution} $\mathcal{F}(\cdot)$. This convolution ensures the multi-scale feature representation $\mathbf{x}'$ for the focal stack images. Subsequently, the \emph{transformer encoder} $g(\cdot)$ processes the feature maps by transforming them into a series of ordered tokens that share information through self-attention. The self-attention weights between the in-focus features and blur features, encode spatial feature information from each input image. The \emph{recurrent LSTM module} sequentially processes cached latent tokens from different frames of a focal stack and fuses them along the stack dimension. The LSTM module learns the stack feature fusion process within the latent space.
Our attention design combined with LSTM enhances the model's capability to handle an arbitrary number of input images. The final disparity map is decoded (denoted as $d(\cdot)$) from the fusion features, utilizing the aggregated tokens from all images in a focal stack.

\begin{figure*}[!thbp]
\centering
\vspace{-0.20em}
\includegraphics[trim=0.6cm 0.3cm 0.8cm 0.6cm, width=0.92\textwidth]{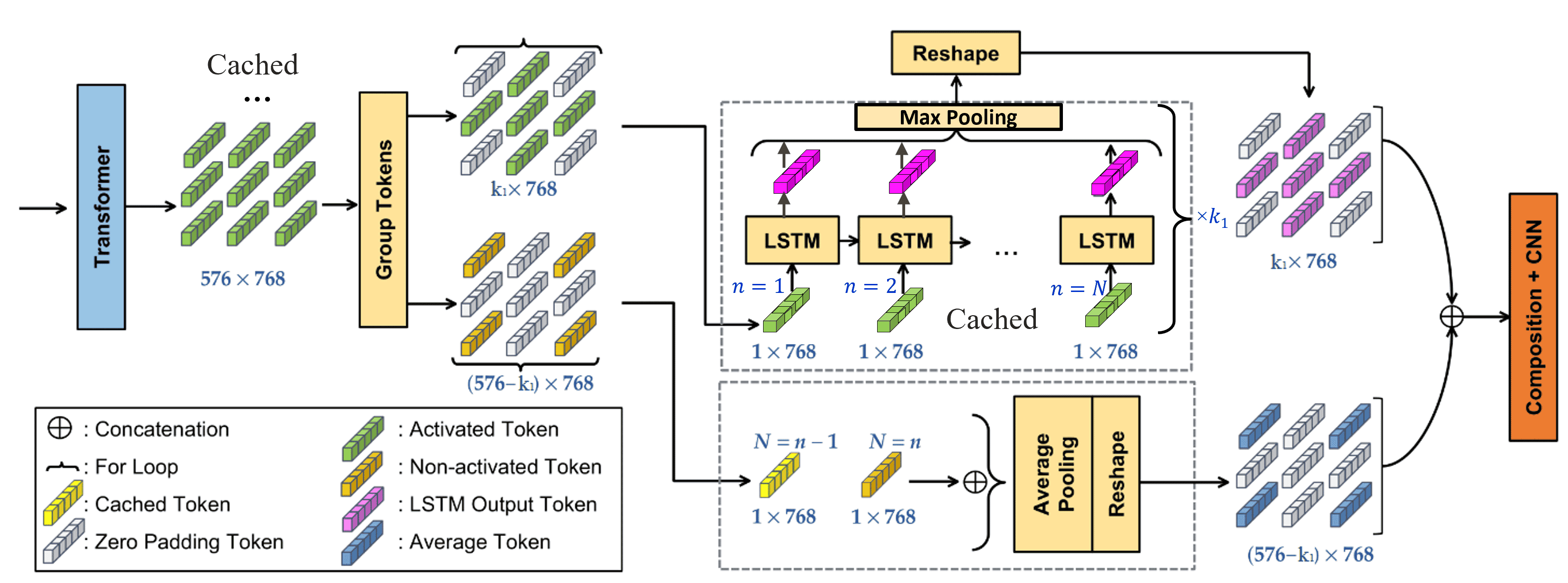}
\vspace{-0.2em}
\caption{To illustrate the LSTM module in our network, the initial step involves grouping the all cached output tokens from the Transformer encoder into activated and non-activated tokens. These two groups are then individually processed, with activated tokens undergoing LSTMs followed by max pooling and non-activated tokens undergoing average pooling. Following this, the output tokens undergo reshaping and concatenation before being fed into the CNN decoder for predicting the depth map.}
\label{fig:LSTM-token}
\end{figure*}

\subsection{Early-stage encoding with multi-scale kernels}
\label{sec:early-CNN}
To capture low-level focus and defocus features at different scales, we employ an early-stage convolutional encoder with multi-scale kernels, which is different from methods using fixed size kernel convolution stem before the Transformer \cite{xiao2021early}. As illustrated in Fig. \ref{fig:model_arch}, the early-stage encoder utilizes three convolutional kernels to generate multi-scale feature maps $f_{m}(\mathbf{x}), \{m = 1,2,3\}$. All feature maps are concatenated and merged into the feature map $\mathbf{x'}\in\mathbf{R}^{H'\times W'\times 1}$ through spatial convolution, followed by $3\times3$ and $1\times1$ convolution on the feature map depth channel:
\begin{equation}
    \mathbf{x'} = \mathcal{F}(\mathbf{x}) = \text{{Conv}}(\text{{Concat}}(f_{m}(\mathbf{x}))),
\end{equation}
where $m$ ranges from $1$ to $3$. Feature concatenation after convolutions with multiple kernel sizes preserves fine-grained details of features across varying depth scales. The first module from the left in Fig. \ref{fig:model_arch}, comprising parallel multi-scale kernel convolutions followed by depth-wise convolution, ensures the model has a large receptive field even beyond the $7\times7$ kernel size. This facilitates capturing more defocus features while preserving intricate details. 
\vspace{-1.2em}
\subsection{Transformer with LSTM}
\begin{figure}
\vspace{-0.2em}
    \centering
    \subfigure[Focus at front.]
    {
          \includegraphics[width=0.22\linewidth]{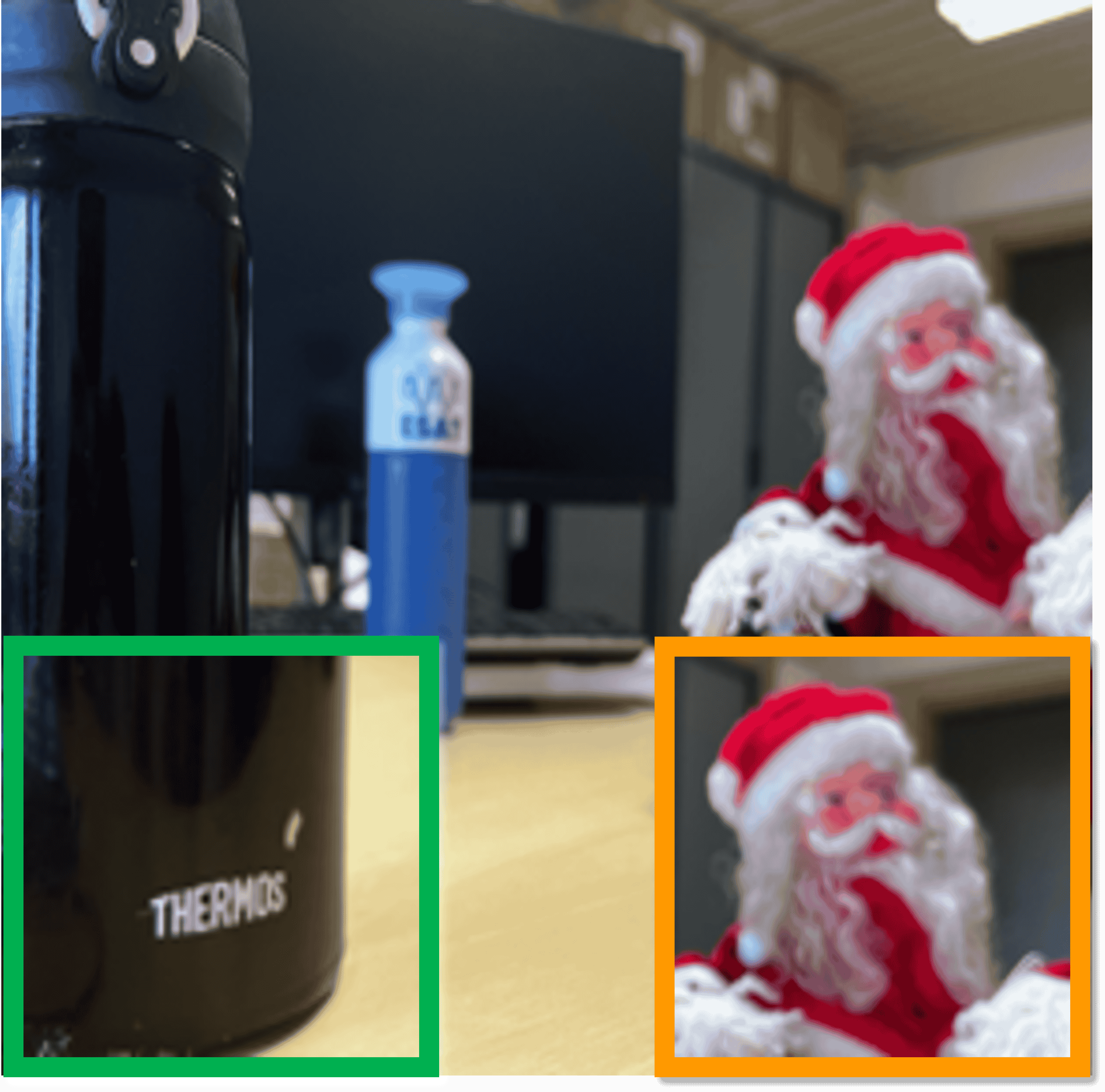}
          \label{fig:rgb-fore}  
    }
    \subfigure[Attention of the (a).]
    {
          \includegraphics[width=0.22\linewidth]{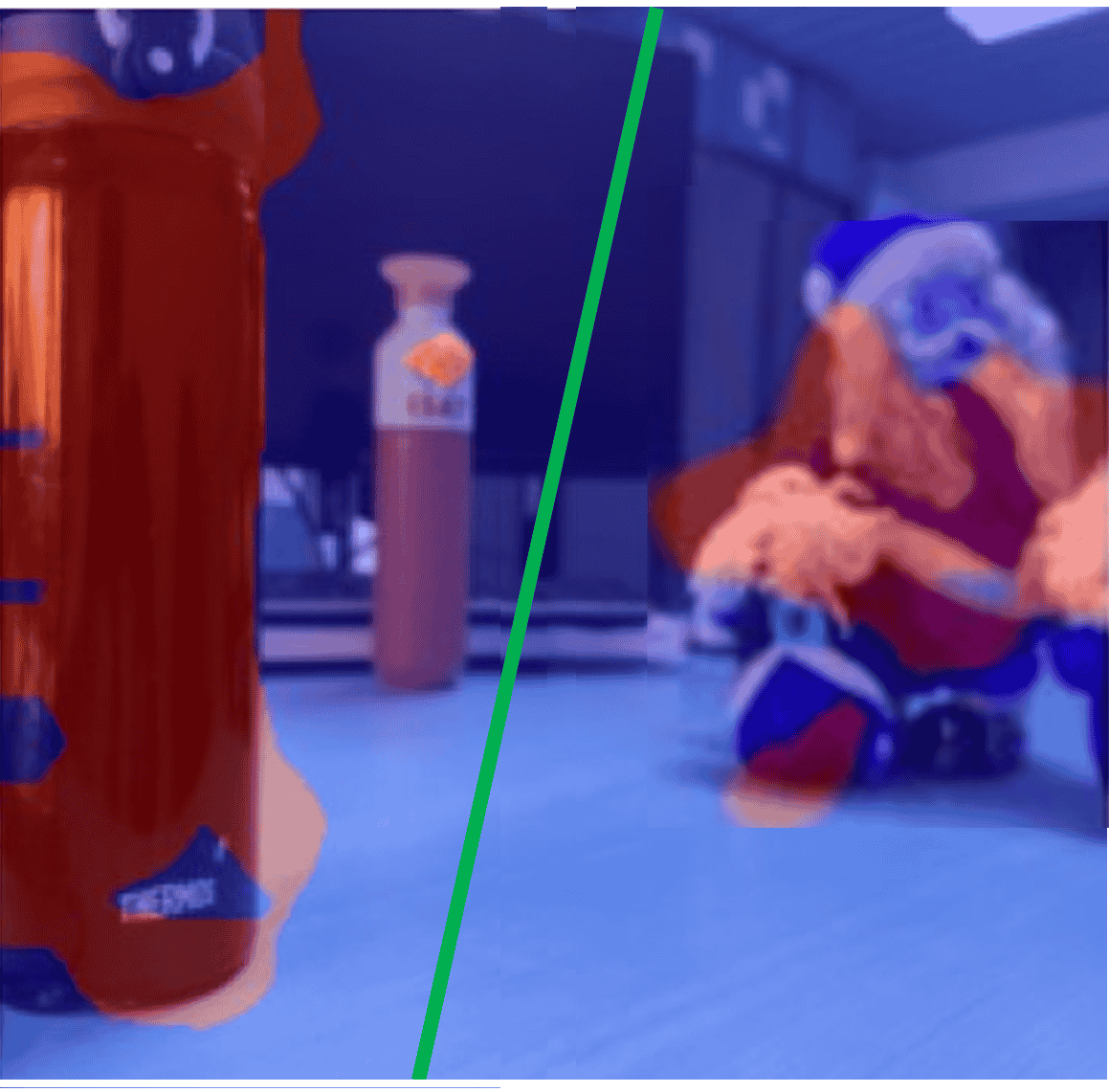}
          \label{fig:atten-fore}  
    }
    \subfigure[Focus at back.]
    {
          \includegraphics[width=0.22\linewidth]{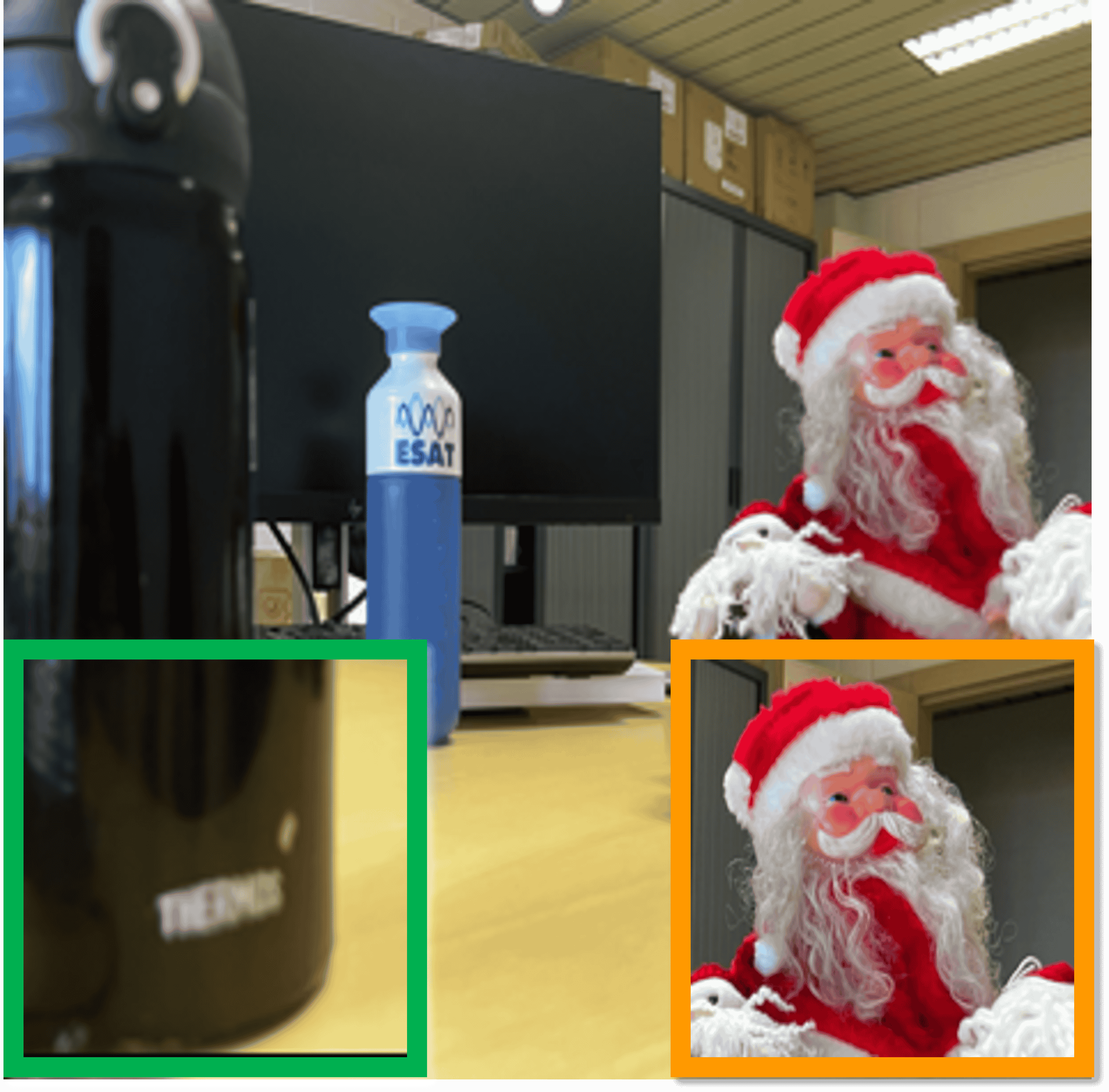}
          \label{fig:rgb-back}  
    }
    \subfigure[Attention of (c).]
    {
          \includegraphics[width=0.22\linewidth]{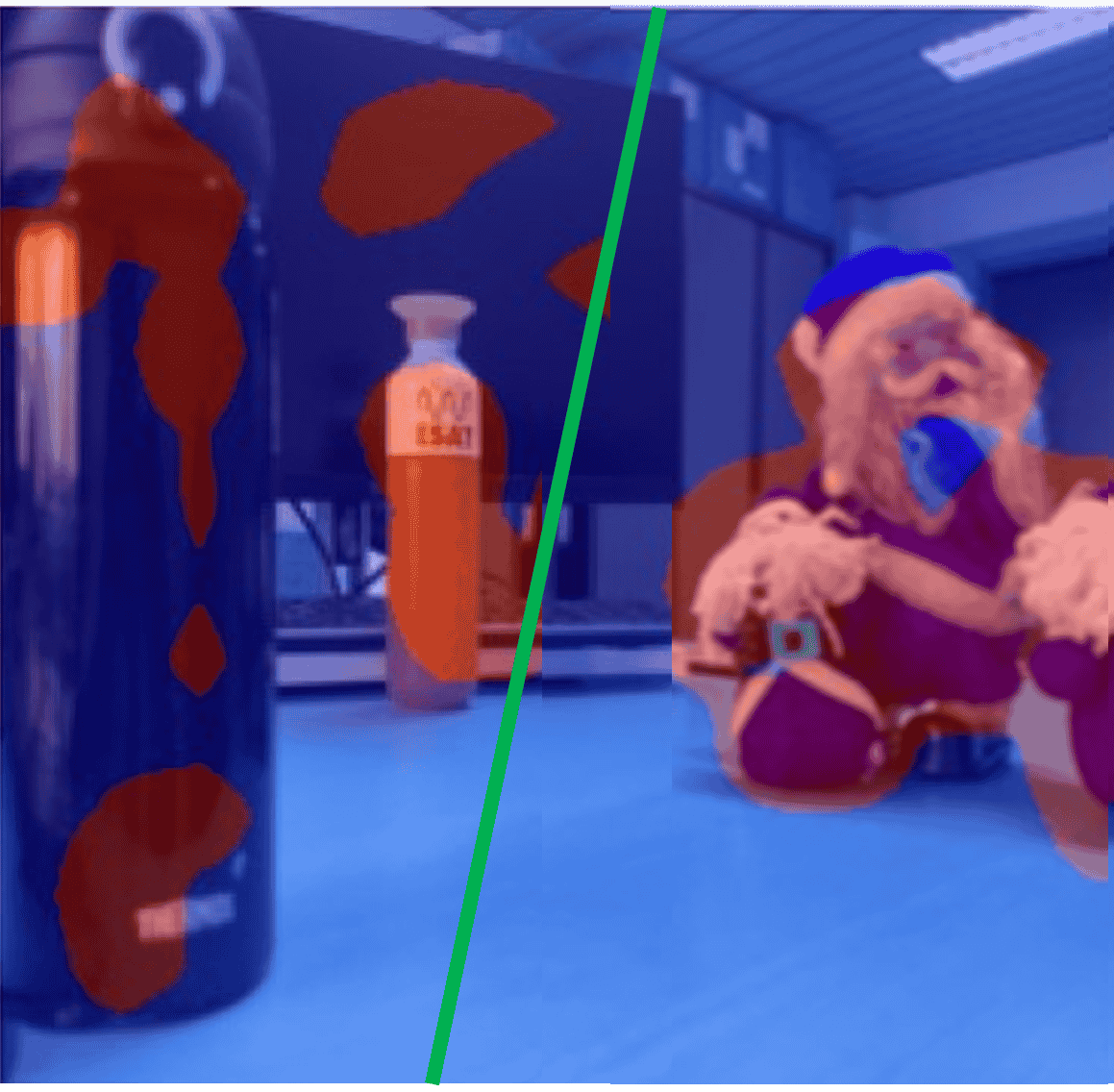}
          \label{fig:atten-back}  
    }
    \caption{Comparison of Transformer attention between the two left column images. Cropped image patches within green and orange boxes in (a) and (c) serve as query inputs to compute the self-attention map over the entire input image, respectively. In (b) and (d), the attention maps on the left and right sides of the green line illustrate the attention outputs of the green and orange boxes, respectively. This demonstrates the model's capability to selectively attend to both foreground and background areas, distinguishing between focus and defocus cues.
}
\label{fig:heatmap-cmps}
\end{figure}

\noindent\textbf{Transformer encoder.} 
The Transformer depicted in Fig. \ref{fig:model_arch}, denoted as $g(\cdot)$, operates on the feature maps $\mathbf{x'}$ derived from preceding early-stage multi-scale convolutions to produce a sequence of tokens $(\mathbf{t'}_{p})_{p=1}^{k}$:
\begin{equation}
    {\mathbf{t'}}_{1}^{}, {\mathbf{t'}}_{2}^{}, ..., {\mathbf{t'}}_{k}^{} = g(\mathbf{x'}).
\end{equation}
Specifically, the early-stage kernel CNNs and Transformer encoder sequentially process the focal stack images, caching and concatenating the feature maps of a specified stack of images into $\mathbf{x'}$. Initially, a linear embedding layer divides the feature maps $\mathbf{x'}$ into $k$ patches of size $16\times 16$. Thus, $\mathbf{x'}_{p} \in \mathbf{x'}, p=1,2,..., k$, is projected by a linear embedding layer (MLP) into corresponding embedding tokens $(\mathbf{l}_{p})_{p=1}^k$, each token having a dimension of $768$ (576 in total). All tokens of a complete stack $N$ are cached into $({\mathbf{l}_{p})_{p=1}^{k}} \times N$ before LSTM for simultaneous fusion. The Transformer's \emph{Position Embedding} encodes the positional information of image patches in a sequential order from the top-left of the image. An MLP layer generates the Global Embedding Token (Fig. \ref{fig:model_arch}) by mapping the entire image into a global token and subsequently adding each individual patch embedding token. Each linear embedding token is projected into three vectors - Query, ${\mathbf{l}}^{}_{Q}$; Key, ${\mathbf{l}}^{}_{K}$; and Value, ${\mathbf{l}}^{}_{V}$ via a weight matrix $\mathbf{W}_{}^{(\cdot)}$ with dimensions $d_Q^{}, d_K^{},$ and $d_V^{}$ respectively. Queries, Keys, and Values are processed in parallel through Multi-Head Attention (MHA) units.
\begin{align}
\vspace{-1.2em}
&\mathrm{MHA}({\mathbf{l}}^{}_{Q},{\mathbf{l}}^{}_{K},{\mathbf{l}}^{}_{V})= ({\mathrm{head}}_1^{} \oplus ... \oplus {\mathrm{head}}_N^{})\mathbf{W}^{O}_{}, \\
&{\mathrm{head}}_i^{}= \mathrm{softmax}\left(\frac{{\mathbf{l}}^{}_{Q}\mathbf{W}_{}^{{\mathbf{l}}^{}_{Q}}{\mathbf{l}}^{}_{K}\mathbf{W}_{}^{{\mathbf{l}}^{}_{K}}}{\sqrt{d_k^{}}}\right){\mathbf{l}}^{}_{V}\mathbf{W}_{}^{{\mathbf{l}}^{}_{V}},
\label{eq:mha}
\vspace{-0.8em}
\end{align}
where $\mathbf{W}_{}^{\mathbf{l}_Q}\in\mathbf{R}^{d_{m}\times d_{V}}$, $\mathbf{W}_{}^{\mathbf{l}_K}\in\mathbf{R}^{d_{m}\times d_{K}}$, $\mathbf{W}_{}^{\mathbf{l}_V}\in\mathbf{R}^{d_{m}\times d_{V}}$, and $\mathbf{W}_{}^{O}\in\mathbf{R}^{d_{m}\times d_{V}}$. Following the Multi-Head-Attention modules within the encoder $g(\mathbf{x'})$, the resulting tokens $(\mathbf{t'}_{p})_{p=1}^{k}$ capture features that distinguish focus and defocus cues among different stack image patches at the same spatial location within the image. This capability is illustrated in Fig. \ref{fig:heatmap-cmps}. Consequently, the embedding space emphasizes sharper, more in-focus features of the image patches.

\noindent\textbf{LSTM module.} To ensure our model's flexibility in handling stacks of arbitrary lengths, in contrast to the fixed lengths used in existing methods \cite{yang2022deep}, we employ an LSTM to progressively integrate sharp features across the stack. The LSTM treats patch embedding tokens at the same image position as sequential features along the stack dimension. Corresponding feature tokens $(\mathbf{t'}_p)_{p=1}^{k}$ from stack images at stack number $N$ are sequentially ordered spatially and fed into LSTM modules arranged in the original spatial image order. At each position, each LSTM module incrementally integrates the latent token $\mathbf{t'}_p$ from a stack. This approach differs from existing models constrained to a 3D volume stack with a predefined and fixed size \cite{yang2022deep}. Importantly, sequential processing in the latent space after a shared encoder for stack images ensures manageable complexity in practice.

\par
Preceding the LSTM modules, tokens associated with image patches from a single frame are classified into activated and non-activated tokens, indicating the level of informativeness of the features. The $L_{2}$ norm, denoted as $\|\cdot\|$, of each embedding token is compared against a threshold of $0.4$, as depicted in Fig. \ref{fig:LSTM-token}. Specifically, within a single frame, only tokens surpassing this threshold are considered activated and forwarded to the LSTM, totaling $k_1$ tokens. This approach significantly reduces the computational load of the LSTM by processing only a subset of the latent tokens:
\begin{flalign}
\vspace{-1.6em}
({\mathbf{t}}^{n}_{p}, \mathbf{h}_{p}^{n}) &= LSTM({\mathbf{t'}}_{p}^{n}, {\mathbf{h}}_{p}^{n-1}, c^{n}),
\label{eq:lstm-group}
\end{flalign}
this is a single LSTM layer expression above, where $p=1,2,...,k_1$, and $n$ is the frame index number of a stack. Here we set the number of hidden layers of the LSTM equal to the stack size $N$.
The memory cell $c^{}$ undergoes continuous updates at each step, influenced by the input $\mathbf{t'}^{n}_{p}$ and the hidden state $\mathbf{h}$. Subsequently, all LSTM layer outputs are combined via max pooling $\max{\mathbf{t'}_{p}^{1}, ..., \mathbf{t'}_{p}^{n}}$ to obtain $\mathbf{t'}_{p}^{}$. For non-activated tokens $(\mathbf{t'}_p), p=k+1,..., k$, an averaging operation is performed with corresponding cached tokens from the previous step at the same embedding position. Finally, the two groups of output tokens are arranged in the original input embedding order, yielding the final fused tokens $(\mathbf{t}_p)_{p=1}^{k}$.

\noindent\textbf{CNN decoder.}Our decoder $d(\cdot)$ adopts the methodology introduced by Ranftl \textit{et al.} \cite{ranftl2021vision}, utilizing Transpose-convolutions to integrate feature maps following LSTMs. Additionally, the decoder incorporates feature maps from the $i$-th layer $g_{i}(\mathbf{x}')$ of the encoder through skip connections, as illustrated in Fig. \ref{fig:model_arch}. Ultimately, decoder $d(\cdot)$ outputs the depth prediction. 
\begin{equation}
    \vspace{1.6em}
    \mathbf{\smash{\hat{D}}} = d((\mathbf{t}_p)_{p=1}^{k}, g_{i}(\mathbf{x}')),\quad i=\{1,2,3\}.
    \label{eq:depth_loss}
    \vspace{-1.5em}
\end{equation}
\subsection{Training loss}
\label{sec:loss}
Our training loss comprises a sum of the Mean Squared Error (MSE) loss, denoted as $\mathcal{L}_{MSE}$, and a sharpness regularizer $\mathcal{L}_{log_{}}$ weighted by $\alpha$:
\begin{equation}
 \mathcal{L}_{total} = \mathcal{L}_{MSE}(\mathbf{\smash{\hat{D}_{}^{}}}, \mathbf{\smash{D_{}^{}}}) + \alpha\mathcal{L}_{\log_{}}(\delta_{\mathbf{\Delta{{\smash{\hat{D}}}_{}^{}}}}, \delta_{\mathbf{\Delta{\smash{D}}_{}^{}}}), 
 \label{eq:total_loss}
\end{equation}
where $\mathbf{\smash{D_{}^{}}}$ represents the ground truth depth, and $\mathbf{\smash{\hat{D}{}^{}}}$ indicates the predicted depth. $\mathbf{\Delta}$ denotes the Laplacian operator applied to the predicted and ground truth depth images, respectively. $\delta$ represents the variance of the depth image. The regularization term is formulated as $\log(\delta_\mathbf{{\Delta{\smash{\hat{D}}}{}^{}}} /\delta_{\mathbf{\smash{\Delta{D}_{}^{}}}})$. Pixel blurriness due to out-of-focus effects can be described by the Circle-of-Confusion (CoC),
\begin{equation}
CoC_{}^{} = \frac{1}{2r}\frac{f_{}^{2}}{N_{}^{}(z_{}^{}-{d}_{f})}\Big |1 -\frac{{d}_{f}}{z}\Big |,
\label{eq:C-radius}
\end{equation}
where $N$ denotes the $f$-number, defined as the ratio of the focal length to the effective aperture diameter, and $r$ is the CMOS pixel size. ${d}_{f}$ denotes the focus distance of the lens, and $z$ represents the distance from the lens to the target object. Generally, the range of $z$ is $[0, \infty]$, though in practice, it is always bounded by lower and upper limits. The model aims to learn a depth map from focus and defocus features of stack images.
\subsection{Pre-training with monocular depth prior}
Focal stack datasets are often limited in size due to the high cost and challenges involved in data collection. To address data scarcity and fully exploit the Transformer's potential, we optionally pre-train the Transformer encoder on widely available monocular depth dataset like NYUv2 \cite{silberman2012indoor}, to enhance spatial representation learning further, yet without pre-training, the model performance is still superior over baseline models, as exhibited in following experiment section.

\section{Experiments}
\label{subsec:setup}
\noindent\textbf{Datasets.} We extensively evaluated our model using four benchmark focal stack datasets: DDFF 12-Scene \cite{hazirbas2018deep}, Mobile Depth \cite{benavides2022phonedepth}, LightField4D \cite{honauer2016dataset}, and FOD500 \cite{yang2022deep} (Synthetic dataset). As Mobile Depth has no depth ground truth, so only visual comparison results are provided. Additionally, our model supports pre-training on the monocular RGB-D dataset NYUv2 \cite{silberman2012indoor}. Specifically, we conducted separate training on DDFF 12-Scene and FOD500 for subsequent experiments, while Mobile Depth and LightField4D were used to assess the model's generalizability directly after pre-training on DDFF 12-Scene. Qualitative and quantitative evaluation on FOD500, and quantitative metric evaluation of LightField4D are provided in the supplementary part. Additionally, the more qualitative results of each dataset are available in the supplementary part, please refer to it. Finally, a comprehensive summary of the evaluation datasets, including their individual properties (real or synthetic), with or without the depth ground truth is presented in Tab. \ref{tab:dataset_summary}, along with defocus cause.
\begin{table}[!thbp]
\centering
\vspace{-0.8em}
\caption{\small{Summary of evaluation datasets.}}
\label{tab:dataset_summary}
\begin{adjustbox}{width=0.70\linewidth}
\begin{tabular}{c c c c}
\toprule
  Dataset  &  Image source & GT type  &  Cause of defocus \\ 
  \hline
  DDFF 12-Scene \cite{hazirbas2018deep}  & Real          & Depth       & Light-field settings \\ 
  \hline
  Mobile Depth \cite{benavides2022phonedepth}  & Real          & ---      & Real\\
  \hline
  LightField4D \cite{honauer2016dataset}  & Real          & Disparity &   Light-field settings \\ 
  \hline
  FOD500 \cite{yang2022deep}    & Synthetic       & Depth       & Synthesis blendering \\ 
  \bottomrule 
\end{tabular}
\end{adjustbox}
\vspace{-0.6em}
\end{table}

\noindent \textbf{Implementation details.} For the evaluation on DDFF 12-Scene, we conducted training experiments using our model with and without pre-training on NYUv2 \cite{silberman2012indoor}, presenting results for both scenarios. We employed a patch size of $16\times16$ and an image size of $384\times384$ for the Transformer. Our network utilizes the Adam optimizer with a learning rate of $1\times10^{-4}$ and a momentum of 0.9. The regularization scalar $\alpha$ in Eq. \eqref{eq:total_loss} is set to 0.2. In terms of hardware configuration, all training and tests below were conducted on a single Nvidia RTX 2070 GPU with 8GB of vRAM.

\noindent \textbf{Metric evaluation.}
In this work, we perform quantitative evaluation using the following metrics: Root Mean Squared Error (RSME), logarithmic Root Mean Squared Error (logRSME), relative absolute error (absRel), relative squared error (sqrRel), Bumpiness (Bump), and accuracy threshold at three levels ($1.25$, $1.25^2$, and $1.25^3$).

\noindent \textbf{Runtime.}
We evaluated the runtime of the proposed method and baseline approaches by executing them on focal stacks from DDFF 12-Scene. Our FocDepthFormer processes a stack of 10 images sequentially in 15ms, averaging 2ms per image. In comparison, DDFFNet \cite{hazirbas2018deep} requires 200ms per stack under the same conditions, and DFVNet \cite{yang2022deep} performs in the range of 20-30ms. 
\begin{table*}[!htbp]
\vspace{-0.4em}
\caption{Evaluation results on DDFF 12-Scene. The best results are denoted in \textbf{\textcolor{red}{Red}} while \underline{\textcolor{blue}{Blue}} indicates the second-best. $\delta = 1.25$.}
\label{tab:ddff-val}
\begin{small}
\begin{adjustbox}{width=\textwidth}
\begin{tabular}{c c c c c c c c c c}
\hline
Model  & RMSE$\downarrow$ & logRMSE$\downarrow$ & absRel$\downarrow$ & sqrRel$\downarrow$ & Bump$\downarrow$& $\delta\uparrow$ & ${\delta}^{2} \uparrow$ & ${\delta}^{3} \uparrow$\\ 
\hline
DDFFNet \cite{hazirbas2018deep} & 2.91e-2 & 0.320 & 0.293 & 1.2e-2 & 0.59 & 61.95 & 85.14 & 92.98\\ 
\hline
DefocusNet \cite{maximov2020focus} & 2.55e-2 & 0.230 & 0.180 & 6.0e-3 & 0.46 & 72.56 & 94.15 & 97.92\\ 
\hline  
DFVNet \cite{yang2022deep} & 2.13e-2 & 0.210 & \underline{\textcolor{blue}{0.171}} & 6.2e-3 & 0.32 & 76.74 & 94.23 & 98.14\\ 
\hline
AiFNet \cite{wang2021bridging} & 2.32e-2 & 0.290 & 0.251 & 8.3e-3 & 0.63 & 68.33 & 87.40 & 93.96\\ 
\hline
Ours (w/o Pre-training)  & \underline{\textcolor{blue}{2.01e-2}} & \underline{\textcolor{blue}{0.206}} & 0.173 & \underline{\textcolor{blue}{5.7e-3}} & \underline{\textcolor{blue}{0.26}} & \underline{\textcolor{blue}{78.01}} & \underline{\textcolor{blue}{95.04}} & \underline{\textcolor{blue}{98.32}}\\ 
\hline
Ours (w/ Pre-training) & \textbf{\textcolor{red}{1. 96e-2}} & \textbf{\textcolor{red}{0.197}} & \textbf{\textcolor{red}{0.161}} & \textbf{\textcolor{red}{5.4e-3}} & \textbf{\textcolor{red}{0.23}} & \textbf{\textcolor{red}{79.06}} & \textbf{\textcolor{red}{96.08}} & \textbf{\textcolor{red}{98.57}}\\ 
\hline  
\end{tabular}
\end{adjustbox}
\end{small}
\vspace{-0.8em}
\end{table*}

\begin{table}[!thbp]
\centering
\caption{Metric evaluation results on "additional" set of LightField4D dataset. The best results are denoted in \textbf{\textcolor{red}{Red}}, while \underline{\textcolor{blue}{Blue}} indicates the second-best.}
\begin{adjustbox}{width=0.70\linewidth}
\centering
\begin{tabular}{ c c c c c c}
  \hline
  Model  & RMSE$\downarrow$ & logRMSE$\downarrow$ & absRel$\downarrow$ & Bump$\downarrow$& $\delta(1.25)\uparrow$\\ 
  \hline
  DDFFNet& 0.431 & 0.790 & 0.761 & 2.93 & 44.39\\ 
  \hline
  DefocusNet & 0.273 & 0.471 & 0.435 & 2.84 & 48.73\\ 
  \hline  
  DFVNet & 0.352 & 0.647 & 0.594 & 2.97 & 43.54\\ 
  \hline
  AiFNet & \textbf{\textcolor{red}{0.231}} & \textbf{\textcolor{red}{0.407}} & \underline{\textcolor{blue}{0.374}} & \underline{\textcolor{blue}{2.53}} & \underline{\textcolor{blue}{55.04}}\\ 
  \hline
  Ours & \underline{\textcolor{blue}{0.237}} & \underline{\textcolor{blue}{0.416}} & \textbf{\textcolor{red}{0.364}} & \textbf{\textcolor{red}{1.54}} & \textbf{\textcolor{red}{58.90}}\\ 
  \hline  
\end{tabular}
\end{adjustbox}
\label{tab:lightfield4d}
\vspace{-0.6em}
\end{table}

\setlength{\tabcolsep}{15pt}
\renewcommand{\arraystretch}{1}
\begin{figure}[!thbp]
\vspace{-0.4em}
\begin{tabularx}{\textwidth}{c c c c  c c c}
\centering
    \hspace{0.2cm} Input & \hspace{0.1cm} GT & \hspace{0.1cm} DDFFNet & \hspace{-0.6cm} DefocusNet & \hspace{-0.2cm}AiFNet & \hspace{-0.0cm}DFVNet & \hspace{0.10cm} Ours  \\
\end{tabularx}
  \includegraphics[width=0.92\textwidth]{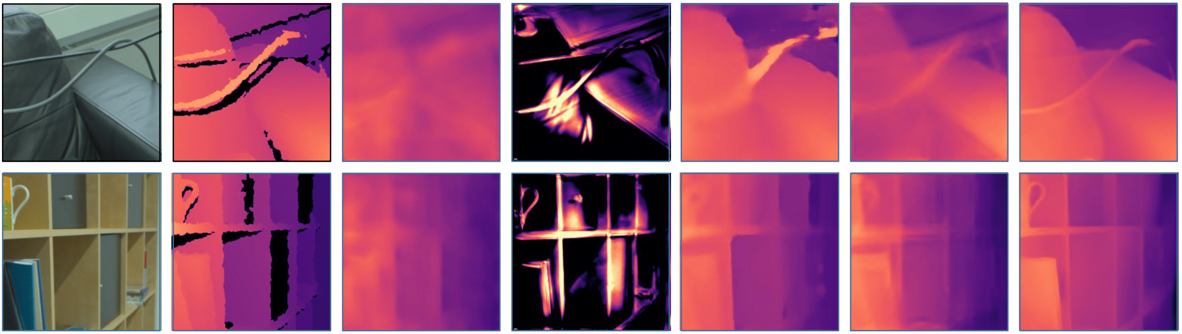}
  \centering
  \vspace{-1.4em}
  \caption{Qualitative evaluation of our model on DDFF 12-Scene dataset.}
  \label{fig:cmp-ddff}
  \vspace{-0.4em}
\end{figure}

\begin{figure}[!thbp]
\vspace{-0.6em}
\begin{tabularx}{\textwidth}{c c  c c  c c c}
\centering
    \hspace{1.6cm} Input & \hspace{-0.2cm} DDFFNet & \hspace{-0.6cm} DefocusNet &\hspace{-0.6cm} AiFNet & \hspace{-0.2cm} DFVNet & \hspace{-0.2cm}Ours  \\
\end{tabularx}
 \centering
    \includegraphics[trim=0cm 8.8cm 0.5cm 0.0cm, width=1.0\textwidth]{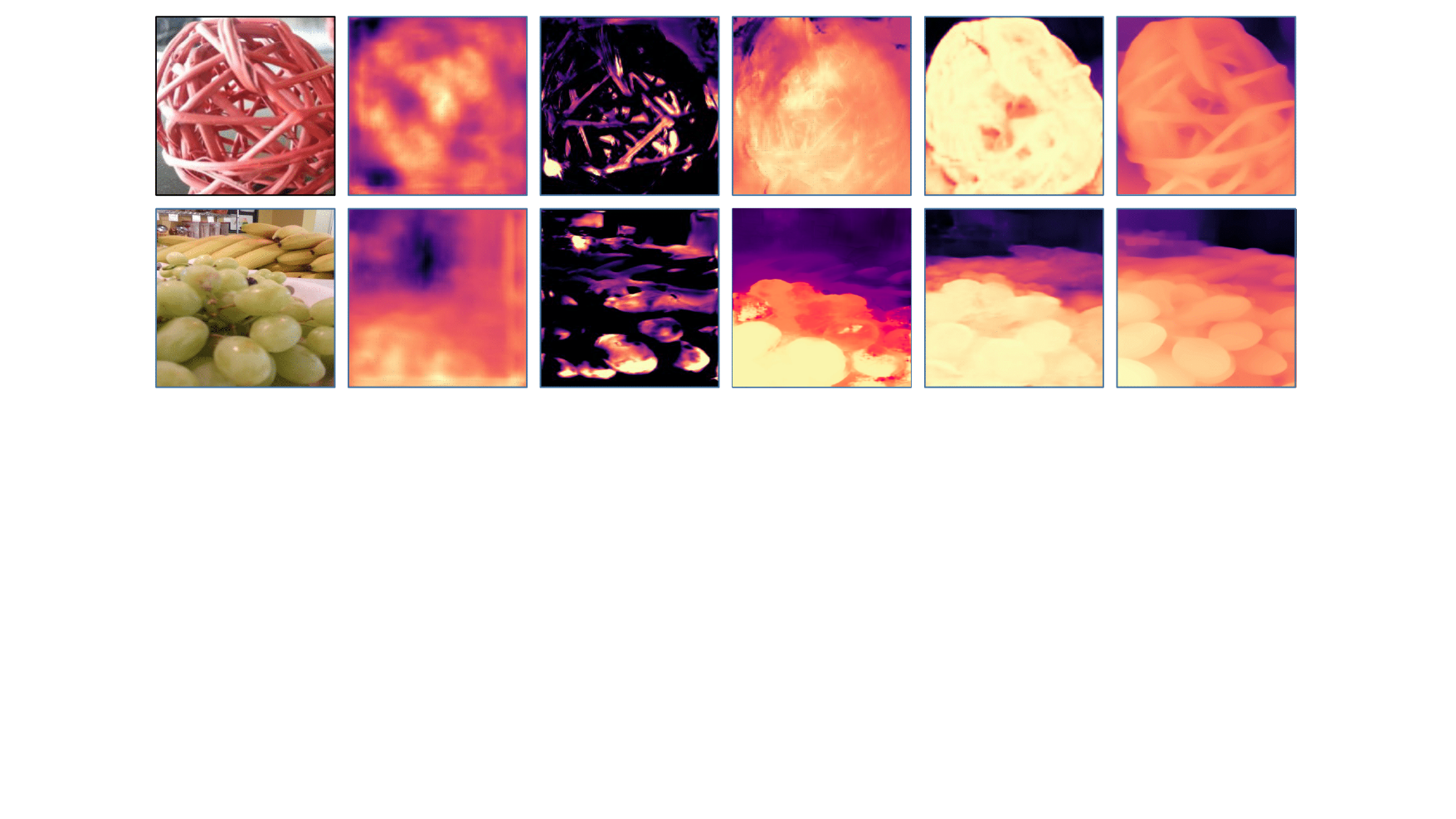}
  \vspace{-1.4em}
  \caption{Qualitative evaluation of our model on Mobile Depth dataset.}
  \label{fig:cmp-mobile}  
  \vspace{-1.0em}
\end{figure}
\subsection{Comparisons to the state-of-the-art methods}
DDFFNet \cite{hazirbas2018deep} and DefocusNet \cite{maximov2020focus} lacked pre-trained weights; therefore, we utilized their open-source codebases to train the networks from scratch. Notably, DefocusNet \cite{maximov2020focus} offers two architectures, and we chose the "PoolAE" architecture due to its consistently good performance for comparison. Conversely, for AiFNet \cite{wang2021bridging} and DFVNet \cite{yang2022deep}, we employed the pre-trained weights provided by the authors to conduct the evaluations. We also provide the visual results of the latest DEReD \cite{si2023fully} model in supplementary part, as the publicly available code is not complete for implementation.

\noindent \textbf{Results on DDFF 12-Scene.} Tab. \ref{tab:ddff-val} presents the quantitative evaluation results of our model on the DDFF 12-Scene dataset. As the ground truth for the "test set" is not publicly available, we adhere to the standard evaluation protocol used in other comparative works, assessing the models on the "validation set" as per the split provided by DDFFNet \cite{hazirbas2018deep}. Moreover, we demonstrate that our model, trained on DDFF-12, performs robustly on other completely unseen datasets, highlighting its generalization ability and mitigating concerns of overfitting. Furthermore, the qualitative results are provided in Fig. \ref{fig:cmp-ddff}.

\begin{figure}[!thbp]
\begin{tabularx}{\textwidth}{c c c c  c c c}
    \hspace{0.5cm} Input & \hspace{0.3cm} GT & \hspace{-0.2cm} DDFFNet & \hspace{-0.5cm} DefocusNet & \hspace{-0.4cm}AiFNet & \hspace{-0.2cm}DFVNet & \hspace{-0.1cm} Ours  \\
\end{tabularx}
  \centering
      \includegraphics[trim=0cm 10.8cm 0.8cm 0cm, width=1.0\textwidth]{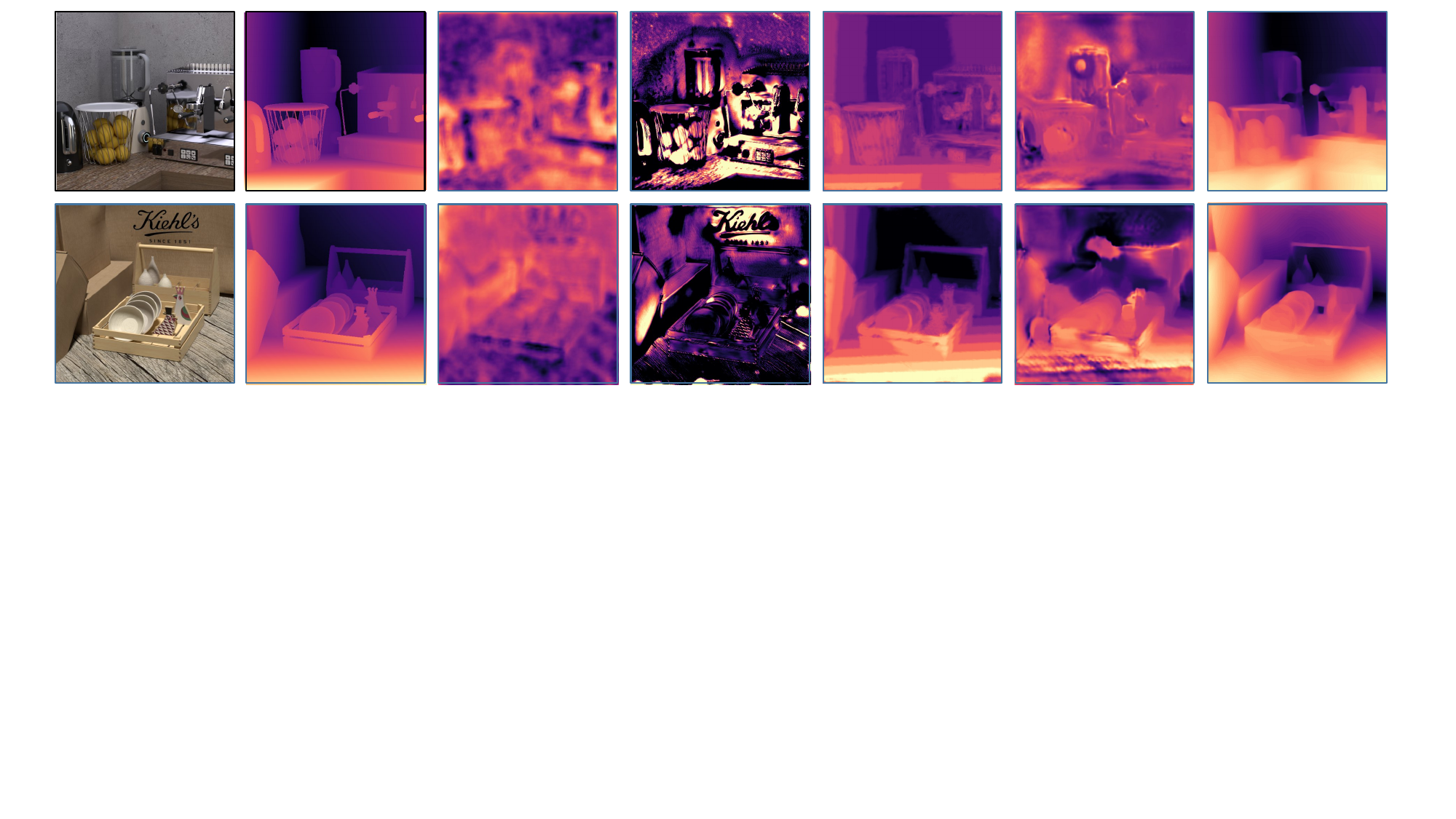}
  \caption{Qualitative evaluation of our model on LightField4D dataset.}
  \label{fig:cmps-lf4d}  
\vspace{-0.2em}
\end{figure}

\noindent\textbf{Results on Mobile Depth, LightField4D, and FOD500.} We provide the qualitative results on Mobile Depth in Fig. \ref{fig:cmp-mobile}, and the depth is not available for this dataset for metric evaluation. Regarding LightField4D, quantitative and qualitative results are provided in Tab. \ref{tab:lightfield4d} and Fig. \ref{fig:cmps-lf4d} respectively. Tab. \ref{tab:ddff-fod} presents the quantitative evaluation of our model on the synthetic FOD500 dtaset and LightField4D. For FOD500, we use the last 100 image stacks from the dataset for test, while the initial 400 image stacks are reserved for training. 
DDFFNet and DefocusNet are re-trained on FOD500 from scratch. The results highlight the consistent superiority of our model across all metrics when compared to the baseline methods. Notably, in our experiments, we observed that pre-training on NYUv2 did not provide significant benefits, likely due to the gap between synthetic and real data. 
\begin{table}[!th]
\vspace{-0.6em}
\centering
\caption{Metric evaluation results on FOD500 test dataset. Here the first 400 FOD500 focal stacks are used for training, following the standard setting from DFVNet \cite{yang2022deep}.
The best results are denoted in \textbf{\textcolor{red}{Red}}, while \underline{\textcolor{blue}{Blue}} indicates the second-best. $\delta = 1.25$.}
\label{tab:ddff-fod}
\begin{small}
\begin{adjustbox}{width=\textwidth}
\begin{tabular}{ c c c c c c c c c c}
  \hline
  Model  & RMSE$\downarrow$ & logRMSE$\downarrow$ & absRel$\downarrow$ & sqrRel$\downarrow$ & Bump$\downarrow$& $\delta\uparrow$ & ${\delta}^{2} \uparrow$ & ${\delta}^{3} \uparrow$ \\ 
  \hline
  DDFFNet \cite{hazirbas2018deep} & 0.167 & 0.271 & 0.172 & 3.56e-2 & 1.74 & 72.82 & 89.96 & 96.26\\ 
  \hline
  
  DefocusNet \cite{maximov2020focus} & 0.134 & 0.243 & 0.150 & 3.59e-2 & 1.57 & 81.14 & 93.31 & 96.62\\ 
  \hline  
  
  DFVNet \cite{yang2022deep} & \underline{\textcolor{blue}{0.129}} & \underline{\textcolor{blue}{0.210}} & \underline{\textcolor{blue}{0.131}} & \underline{\textcolor{blue}{2.39e-2}} & \underline{\textcolor{blue}{1.44}} & 81.90 & \underline{\textcolor{blue}{94.68}} & \underline{\textcolor{blue}{98.05}} \\ 
  \hline
  
  AiFNet \cite{wang2021bridging} & 0.265 & 0.451 & 0.400 & 4.32e-1 & 2.13 & \underline{\textcolor{blue}{85.12}} & 91.11 & 93.12\\ 
  \hline
  
  
  Ours (w/o Pre-training) & \textbf{\textcolor{red}{0.121}} & \textbf{\textcolor{red}{0.203}} & \textbf{\textcolor{red}{0.129}} & \textbf{\textcolor{red}{2.36e-2}} & \textbf{\textcolor{red}{1.38}} & \textbf{\textcolor{red}{85.47}} & \textbf{\textcolor{red}{94.75}} &  \textbf{\textcolor{red}{98.13}}\\ 
  \hline  
\end{tabular}
\end{adjustbox}
\end{small}
\vspace{-1.2em}
\end{table}

We present the quantitative results for cross-dataset evaluation of our model on the LightField4D dataset in Tab. \ref{tab:lightfield4d}. Our model achieves a comparable performance in terms of accuracy ($58.90$\%) on this completely unseen dataset. 

\subsection{Cross dataset evaluation}
To evaluate the generalizability of our model, it is initially trained on DDFF 12-Scene and subsequently evaluated on the Mobile Depth and LightField4D datasets. The Mobile Depth dataset poses a challenge as it comprises 11 aligned focal stacks captured by a mobile phone camera, each with varying numbers of focal planes and lacking ground truth. Results on the Mobile Depth dataset are illustrated in Fig. \ref{fig:cmp-mobile}, showcasing the model's ability to preserve sharp information for depth prediction in complex scenes. 
\vspace{-0.6em}
\subsection{Ablation study}
\label{ablation}
\vspace{-0.6em}
We conduct a comprehensive ablation study to evaluate the key components of our proposed model architecture. Additional ablation experiments are detailed in the supplementary materials.
\par
\vspace{-0.0em}
\noindent\textbf{Multi-scale early-stage kernels:} Tab. \ref{tab:ddff-kernel-ablation} presents a comparison of different early CNN kernel design configurations in conjunction with our Transformer encoder (ViT). The effectiveness of the proposed \emph{early-stage multi-scale kernels} encoder is evident, demonstrating robust performance. Conversely, omitting multi-scale kernels or forgoing subsequent convolutions after in-parallel convolutions leads to a degradation in model performance.
\begin{table}[!thbp]
\centering
\vspace{-0.0em}
\caption{Results of different designs of the early-stage convolution. 
}
\vspace{-0.0em}
\label{tab:ddff-kernel-ablation}
\begin{adjustbox}{width=0.92\linewidth}
\begin{tabular}{c c c c}
\hline
  & RMSE$\downarrow$ & absRel$\downarrow$ & Bump$\downarrow$ \\ 
  \hline
 \emph{early kernel size at $3\times3$} & 2.18e-2 & 0.216 & 0.31 \\ 
  \hline
 \emph{early kernel size at $5\times5$} & 2.20e-2 & 0.214 & 0.32 \\ 
  \hline
 \emph{early kernel size at $7\times7$} & 2.13e-2 & 0.192 & 0.35 \\ 
  \hline 
  \emph{multi-scale combination kernels} & \textbf{2.01e-2} & \textbf{0.173} & \textbf{0.26} \\ 
  \hline 
\end{tabular}
\end{adjustbox}
\vspace{-0.0em}
\end{table}
\par
\begin{figure}[!th]
\vspace{-0.8em}
    \centering
    \subfigure[Ours]
    {
          \includegraphics[width=0.24\linewidth]
          {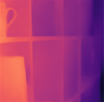}
          \vspace{-0.6em}
          \label{fig:ours-lstm}  
    }
    \subfigure[LSTM+CNN]
    {
          \includegraphics[width=0.254\linewidth]{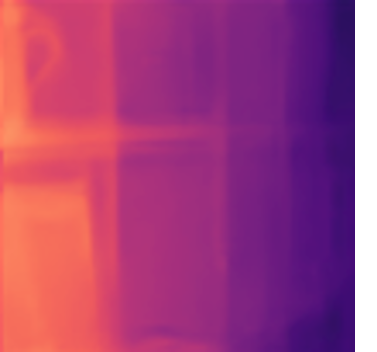}
          \vspace{-0.6em}
          \label{fig:cnn=lstm}  
    } \hspace{-0.4em}
    \subfigure[Transformer]
    {
          \includegraphics[width=0.24\linewidth]{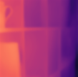}
          \vspace{-0.6em}
          \label{fig:trans}  
    }
    \vspace{-0.45cm}
    \label{fig:cmps-lstm-tf}
    \caption{Different model structure design comparisons.}
    \vspace{-0.0em}
\end{figure}

\noindent\textbf{LSTM for handling arbitrary stack length:} Our proposed LSTM-based method exhibits flexibility in processing focal stacks of arbitrary lengths, a feature distinguishing it from designs that limit inputs to fixed lengths. To illustrate the advantages of our LSTM-based model, we conducted experiments using DDFF 12-Scene. The results, including the RMSE comparison between our model and DFVNet \cite{yang2022deep}, are presented in Tab. \ref{tab:dfv-ours-cmp}. Initially, we trained our model and DFVNet using 10-frame (10F) stacks (\textit{i.e.,} Ours-10F and DFVNet-10F). During testing, DFVNet-10F is constrained and cannot process stacks with fewer than 10 frames. Due to fixed stack size requirements during training and testing, DFVNet must be retrained for different stack sizes (DFVNet-\#F). In contrast, our model is trained once on 10-frame stacks and can be tested with varying numbers of stack images. The focal stack images are ordered based on focus distances. Despite our model's initial performance being inferior to DFVNet, the learning curve of LSTM indicates rapid convergence (Fig. \ref{fig:cmps-lstm-tf}).

\begin{table}[!thbp]
\centering
\vspace{-0.0em}
\caption{RMSE for evaluation of LSTM compared to DFVNet.}
\vspace{-0.0em}
\label{tab:dfv-ours-cmp}
\begin{adjustbox}{width=0.94\linewidth}
\begin{tabular}{ c c c c c c}
  \hline
  Model & 2 Frames & 4 Frames & 6 Frames & 8  Frames & 10 Frames\\ 
  \hline
  Ours-10F & 3.2e-2 & \textbf{2. 61e-2} & \textbf{2.18e-2} & \textbf{2.16e-2} & \textbf{2.04e-2}\\
  \hline 
  DFVNet-10F & ---- & ---- & ---- & ---- & 2.43e-2 \\
  \hline  
  DFVNet-\#F & \textbf{2.97e-2} & 2.70e-2 & 2.52e-2 & 2.47e-2 & 2.43e-2 \\
  \hline    
  \end{tabular}
\end{adjustbox}
\centering
\vspace{-0.0em}
\vspace{-0.0em}
\end{table}

\begin{figure}[!th]
\vspace{-0.0em}
\centering
    \subfigure[1st image.]
    {
          \includegraphics[width=0.30\linewidth]{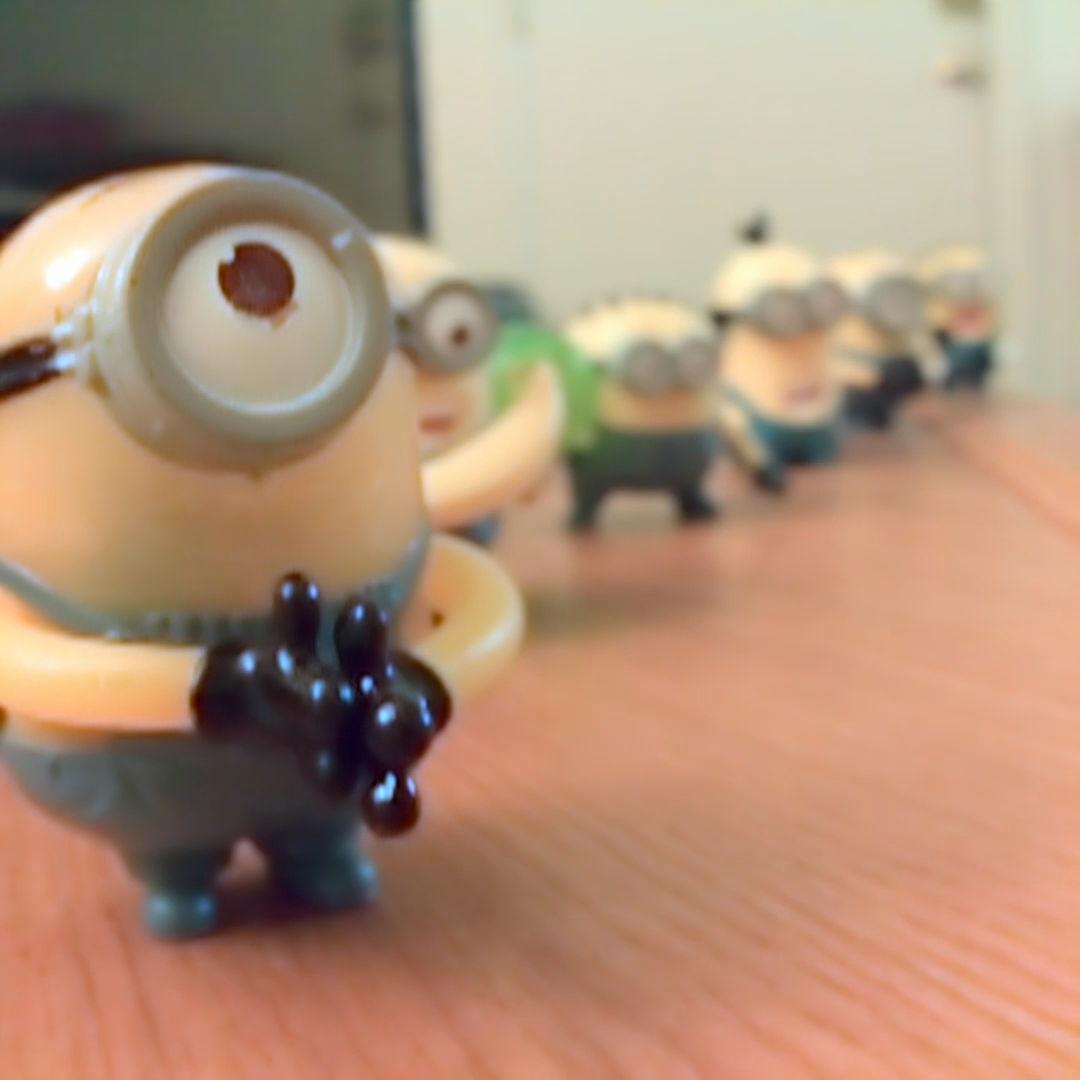}
          \label{fig:rgb-1st}  
    }   
    \subfigure[5th image.]
    {
          \includegraphics[width=0.30\linewidth]{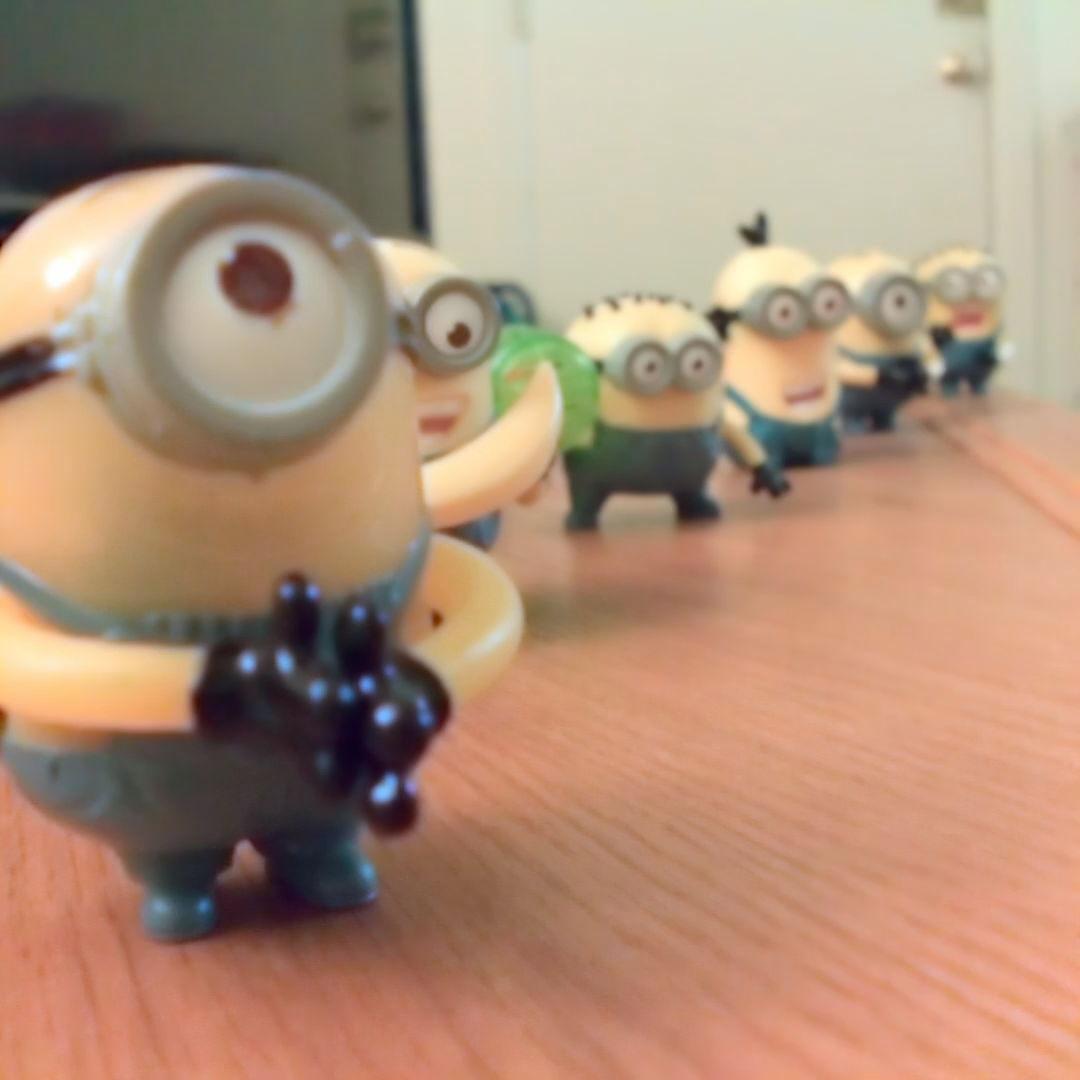}
          \label{fig:rgb-5th}  
    }   
    \subfigure[10th image.]
    {
          \includegraphics[width=0.30\linewidth]{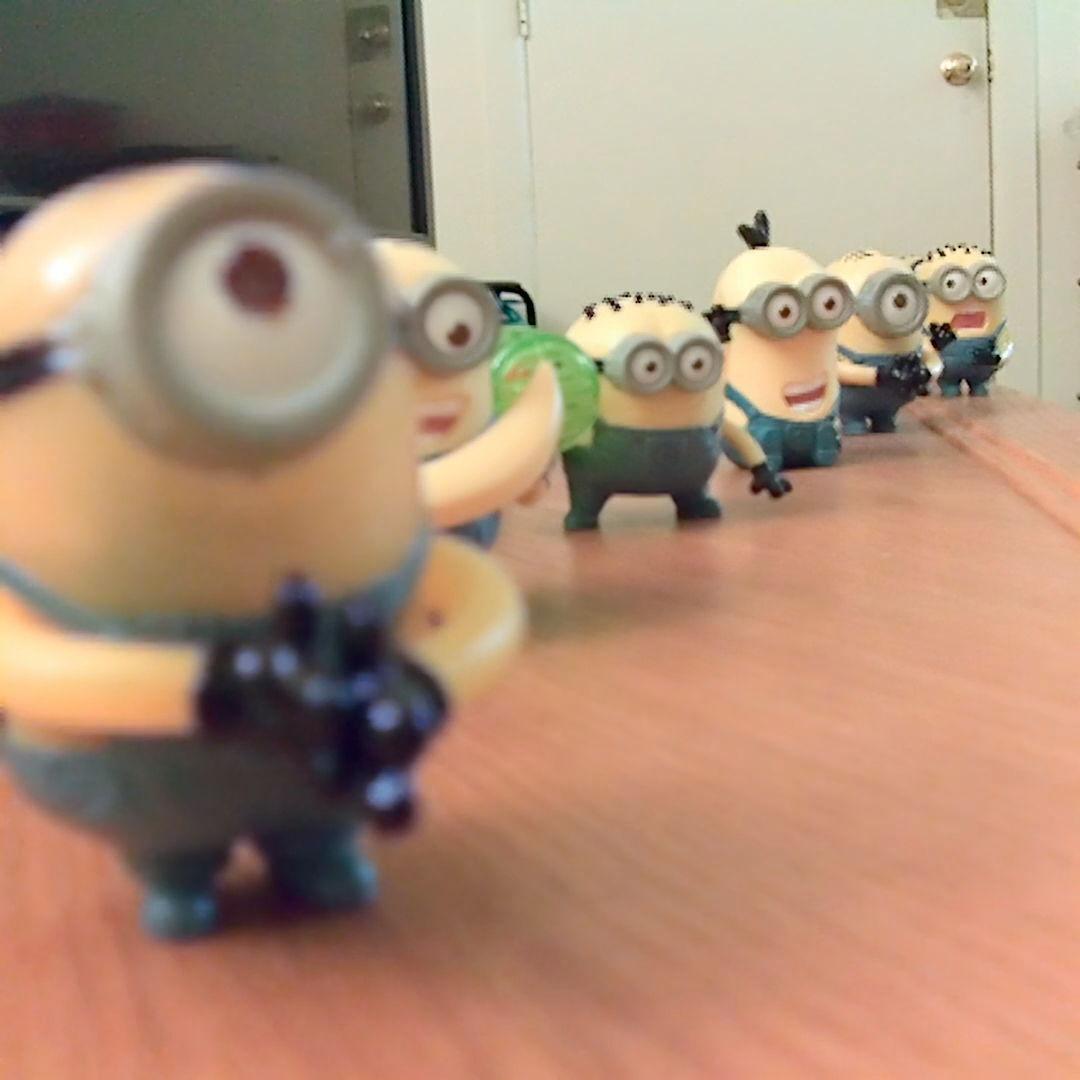}
          \label{fig:rgb-10th}  
    } 
    \subfigure[Depth prediction of 1st input frame.]
    {
          \includegraphics[width=0.30\linewidth]
          {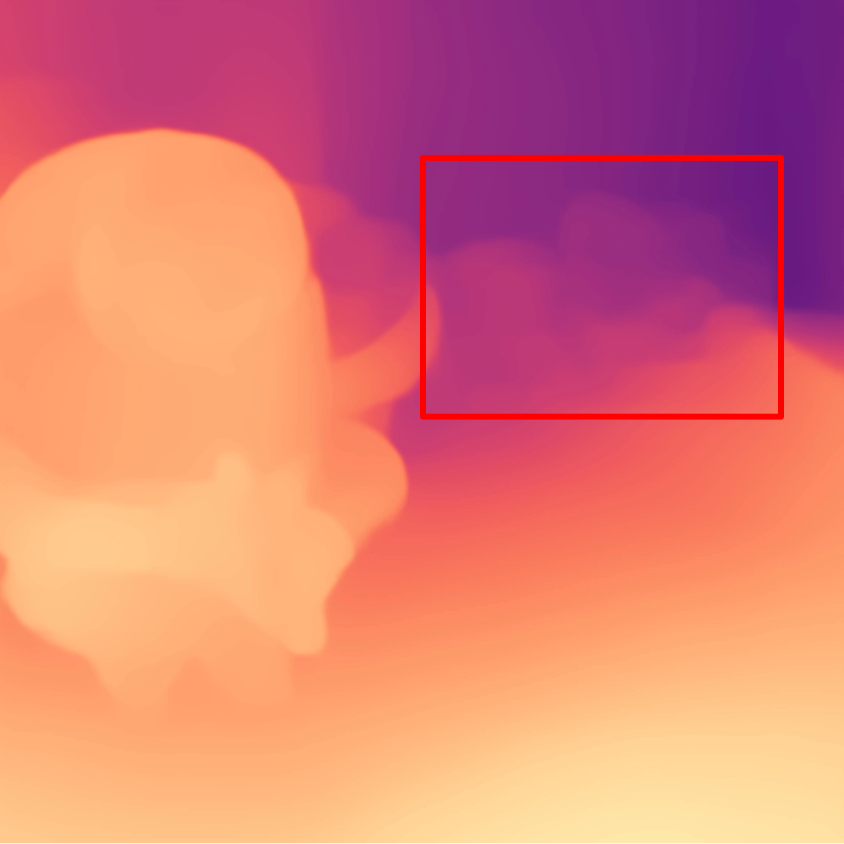}
          \label{fig:depth-1st}  
    } 
      \subfigure[Fusion prediction of 2 input frames.]
    {
          \includegraphics[width=0.30\linewidth]{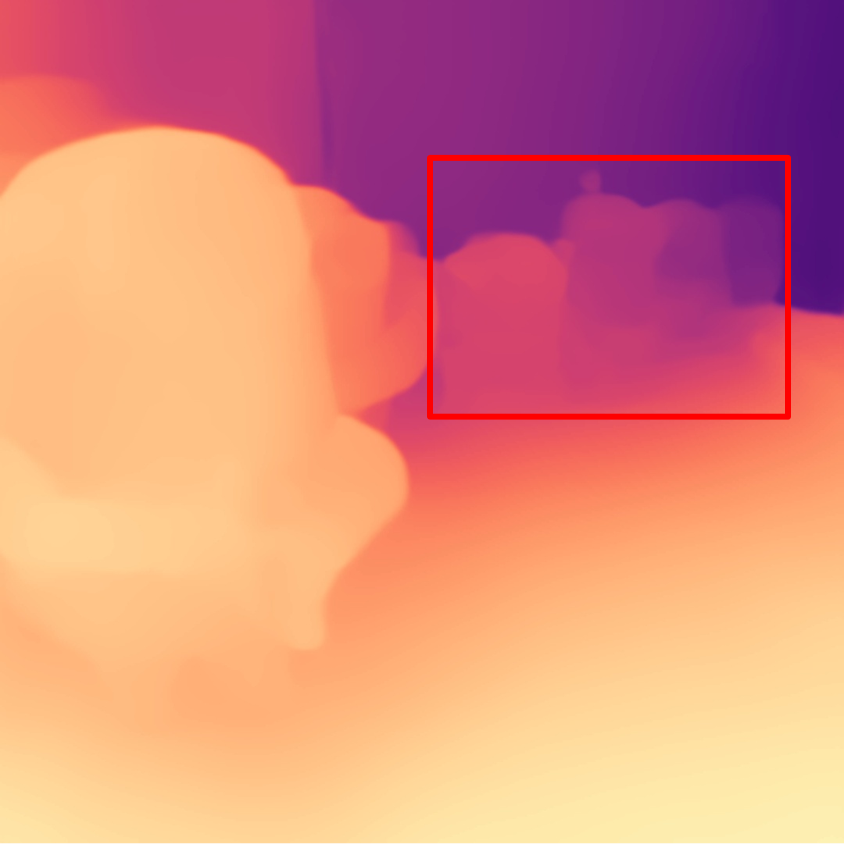}
          \label{fig:depth-2nd}  
    } 
      \subfigure[Fusion prediction of 3 input frames.]
    {
          \includegraphics[width=0.30\linewidth]{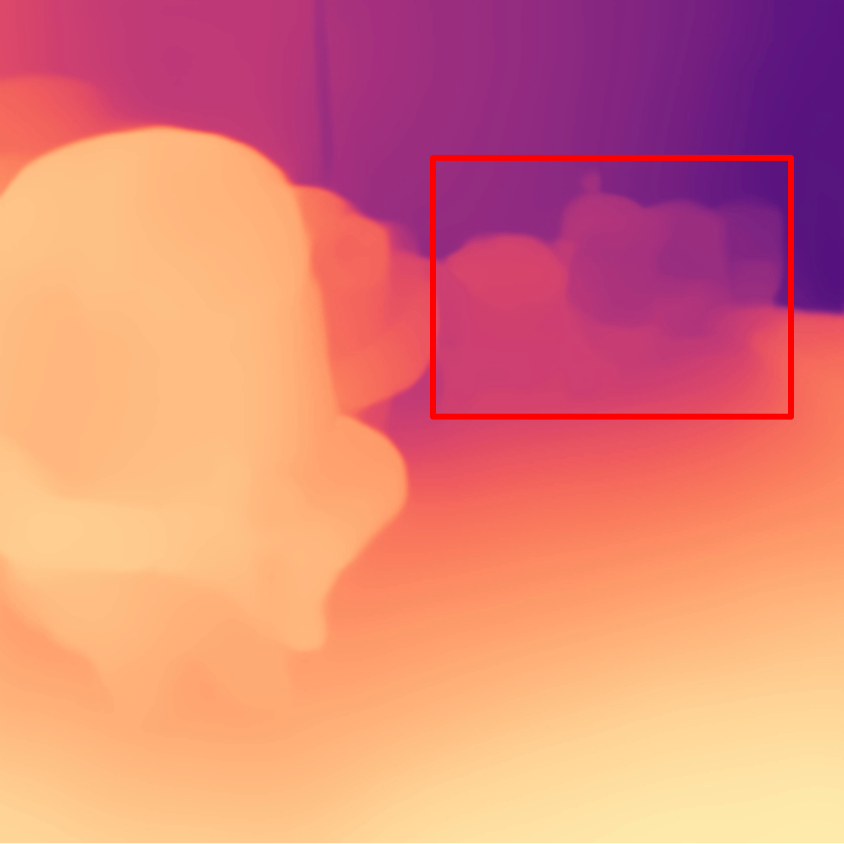}
          \label{fig:depth-3Rd}  
    } 
    \vspace{-0.0em}
\caption{The top row is the input, and the bottom is the output disparity map. The final disparity map is at the bottom right. The red rectangle highlights the incremental fusion results of depth in the background.}
\label{fig:LSTM-image}
\vspace{-0.0em}
\end{figure}

The LSTM module enables our network to incrementally fuse each image from the focal stack, enhancing the model ability to accommodate varying focal stack lengths. Fig. \ref{fig:LSTM-image} depicts the fusion process of the ordered input images of one stack. As the images from the stack are given sequentially, starting from the in-focus plane close to the camera, the model can fuse the sharp in-focus features from a sweep of various frames to attain a final all-in-focus prediction depth map at the bottom right, form near to far focus distances.

Our network comprises the early-stage multi-scale CNN, the Transformer, the LSTM module, and the decoder. We present a summary of each module parameter size, and the inference time,
\begin{table}[!thbp]
\vspace{-0.0em}
\centering
\caption{Summary of module size, and inference time for processing one image, for the whole stack, the total time is calculated from a whole stack of images' processing.}
\vspace{-0.0em}
\begin{adjustbox}{width=0.82\linewidth}
\centering
\begin{tabular}{ c c c }
  \toprule
  Module & Params size &  Inference Time \\ %
  \hline
  Early multi-scale kernel CNNs & 0.487M & 0.001s \\ 
  \hline
  Transformer & 42.065M  & 0.006s \\ %
  \hline
  LSTM & 15.247M & 0.003s\\ %
  \hline   
 CNN Decoder & 16.856M & 0.005s\\ 
  \hline
  \hline
  Total & 74.655M & 0.015\\  
  \bottomrule
\end{tabular}
\end{adjustbox}
\label{tab:model_params}
\vspace{-0.0em}
\end{table}
From Tab. \ref{tab:model_params}, we can see the main time consumption is allocated on the Transformer module and decoder, which shows the potential to reduce the model size further can be achieved, \eg, by using MobileViT as an encoder. Our proposed LSTM over the latent feature representation, has a size of only around one-third of Transformer encoder, furthermore, the shallow early multi-scale kernel CNNs is in quite a small size with only 0.487 million parameters and fast processing time around 1ms. The time model size and time complexity summary table further justify our model's compact and efficient design, where the recurrent LSTM module has benefits both in memory size and computational complexity in the design. The main inference time for a single image processing is from the Transformer Encoder, which can be attributed mainly to the attention computation of multiple self-attention heads, and CNN decoder. 

\noindent\textbf{Loss Function:} Tab. \ref{tab:ddff-loss-ablation} presents the results obtained using three distinct loss functions: Mean Squared Error (MSE), Mean Absolute Error (MAE), and Regularized MSE (MSE with gradient regularization). The findings consistently show that MSE outperforms MAE in terms of overall performance. Notably, incorporating the gradient regularizer contributes to achieving the highest accuracy, as evidenced by the bumpiness metric ($0.26$).

\begin{table}[!thbp]
\centering
\vspace{-1.0em}
\caption{Evaluation for our model with different losses.}
\label{tab:ddff-loss-ablation}
\begin{adjustbox}{width=0.82\linewidth}
\begin{tabular}{c c c c}
\hline
  & RMSE$\downarrow$ & absRel$\downarrow$ & Bump$\downarrow$ \\ 
  \hline
  MSE loss &2.94e-2 & 0.280 & 0.50  \\ 
  \hline
  MAE loss &3.76e-2 & 0.372 & 0.62\\ 
  \hline
  MSE + Gradient loss  & \bf{2.01e-2} & \bf{0.173} & \bf{0.26}  \\ 
  \hline 
\end{tabular}
\end{adjustbox}
\vspace{-0.0em}
\end{table}
\noindent\textbf{Pre-training:} Tab. \ref{tab:pre-train} illustrates the impact of pre-training on the performance of our proposed method. The results demonstrate that pre-training enhances the model's capabilities, leveraging the compact design with Transformer and LSTM modules. Even without pre-training, our model achieves competitive results. Notably, attempts to apply pre-training to DFVNet using a stack created from repeated monocular images of NYUv2 did not yield improvements and, in some cases, led to performance degradation due to the data modality gap.
\begin{table}[!thbp]
\vspace{-1.0em}
\centering
\caption{Pre-training contribution comparisons.}
\label{tab:pre-train}
\begin{adjustbox}{width=\linewidth}
\begin{tabular}{c c c c c c}
\hline
    & Pre-training & RMSE$\downarrow$ & logRMSE$\downarrow$ & absRel$\downarrow$ & Bump$\downarrow$\\%
\hline
    \multirow{2}{*}{Ours} & \xmark & 2.01e-2 & 0.206 & 0.173 & 0.26   \\ 
         & \cmark & \textbf{1.96e-2} &\textbf{0.197} & \textbf{0.161} & \textbf{0.19} \\ 
\hline
    \multirow{2}{*}{DFVNet} & \xmark & 2.13e-2 & 0.210 & 0.171 & 0.32\\
    & \cmark & 2.57e-2 & 0.233 & 0.184 & 0.49\\
\hline
\end{tabular}
\end{adjustbox}
\end{table}
\vspace{-1.0em}
\section{Conclusion}
\vspace{-0.0em}
\label{sec:conclusion}
We introduce the FocDepthFormer model tailored for depth estimation from focal stack. At the core of our network is a Transformer encoder coupled with a recurrent LSTM module in the latent space, allowing the model to effectively capture spatial and stack information independently. Our approach demonstrates flexibility in accommodating varying focal stack lengths. A significant drawback lies in the heightened model complexity associated with the vanilla Transformer architecture. More efficient attention design techniques like Mamba can be explored as the future work. %
%
%
%

\end{document}


\title{Supplementary Material for FocDepthFormer: Transformer with Latent LSTM for \\ Depth Estimation from Focal Stack}
\author{Xueyang Kang\inst{1,2} \and
Fengze Han\inst{3} \and
Abdur R. Fayjie\inst{1} \and
Patrick Vandewalle\inst{1} \and \\
Kourosh Khoshelham\inst{2} \and
Dong Gong\inst{4}
}
\authorrunning{X. Kang et al.}
%
\institute{
Department of ESAT, KU Leuven, Belgium \and
Faculty of Engineering and IT, The University of Melbourne, Australia \and
EI Faculty, Technical University of Munich, Germany \and
EI Faculty, The University of New South Wales, Australia
}
\maketitle

\section{Method}
\subsection{Focal principles}
\label{sec:focal_principle}
In the paper we present the features of our network model, a primary factor to explain our favorable results is the good identification ability of pixel sharpness, as mentioned in our paper. In general, the sharpness of pixels can be evaluated according to the Circle-of-Confusion (CoC) metric, denoted by $C$, defined as below,
\begin{equation}
CoC_{}^{} =
\frac{f_{}^{2}}{N_{}^{}(z_{}^{}-{d}_{f})}\Big |1 -\frac{{d}_{f}}{z}\Big |,
\label{eq:C-radius}
\end{equation}
where, $N$ denotes $f$-number, as a ratio of focal length to the valid aperture diameter. ${d}_{f}$ is the focus distance of the lens. $z$ represents the distance from the lens to the target object. Without loss of generality, the range of $z$ is $[0, \infty]$. However, in reality, the range is always constrained by lower and upper bounds.
\begin{figure}[!thbp]
\centering
\includegraphics[page=1, trim=2.2cm 5.5cm 0.5cm 0.0cm, width=0.5\textwidth]{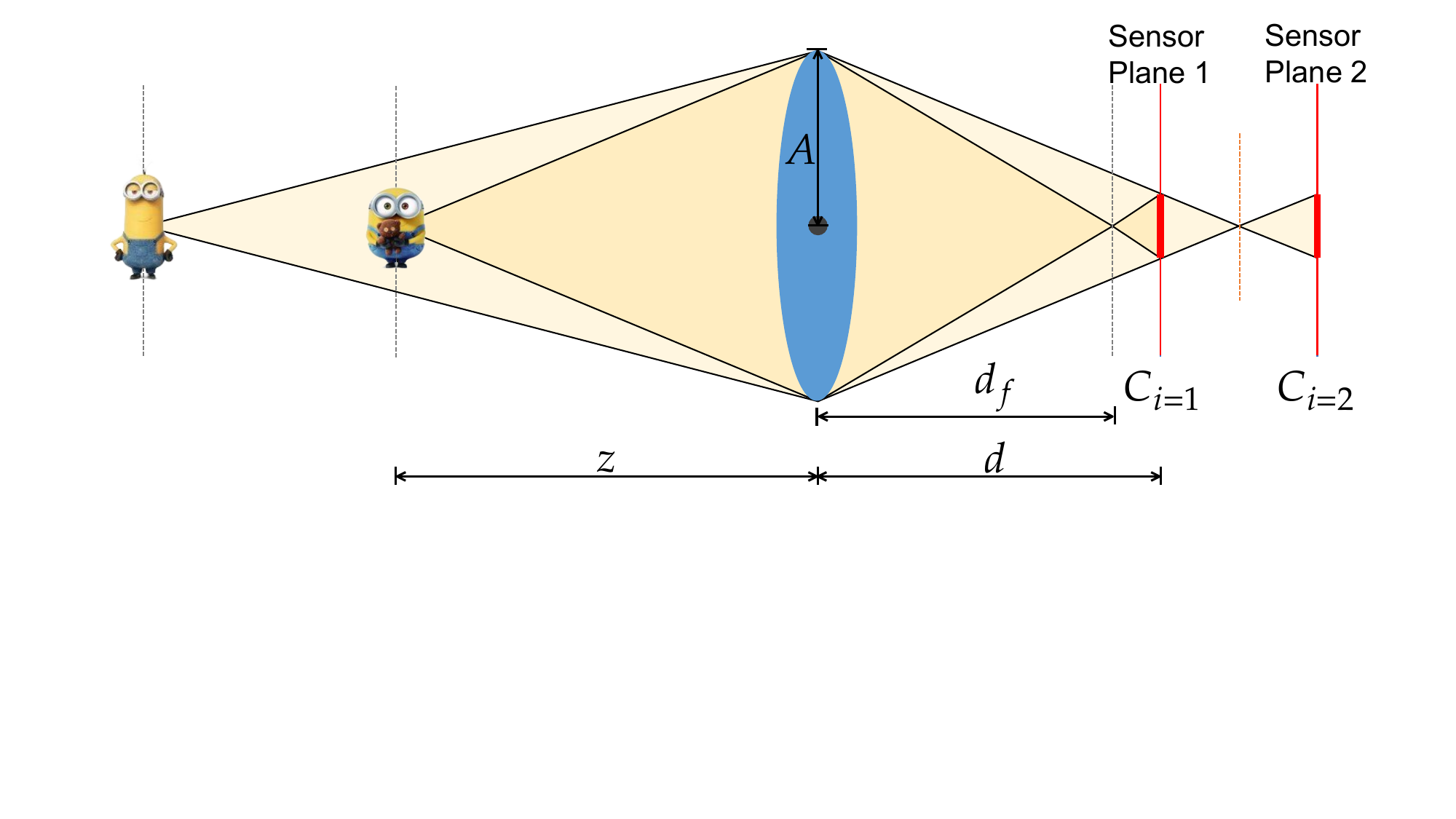}
\caption{The rays emitted from an object, placed at an axial distance $z$ to the lens, converges at a distance ${d}_{f}$ behind the lens. Sensor is situated at a distance $d$ from focal lens. The pixels are sharply imaged when the sensor is placed right at focus distance, $C$ is the CoC, that grows as the sensor position is deviated from the focus plane.}
\label{fig:lens}  
\end{figure} 

In Eq. \ref{eq:C-radius}, $C$ is zero ($C_{}^{*}$) when the image pixel is in-focus, and it is a signed value, where $C>0$ indicates the camera focused in front of the sensor plane, while $C<0$ is the reverse case, with camera focused behind the sensor plane. It can be inferred from the denominator, CoC $C$ has two divergence points, at 0 and $d_{f}$ respectively. Our model actually learn the mapping relationship from CoC to depth in an internally driven manner.

\subsection{Attention map}

\begin{figure}
    \centering
    \subfigure[Attention map of the left patch in Subfig. 3(a) of the paper.]
    {
          \includegraphics[width=0.45\linewidth]{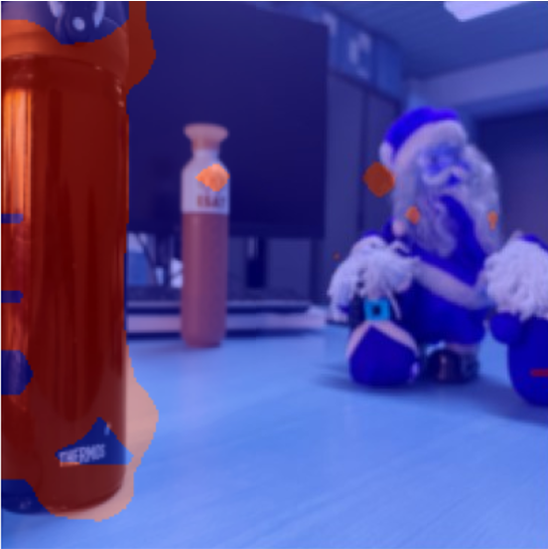}
    }
    \subfigure[Attention map of the right patch in Subfig. 3(a) of the paper.]
    {
          \includegraphics[width=0.45\linewidth]{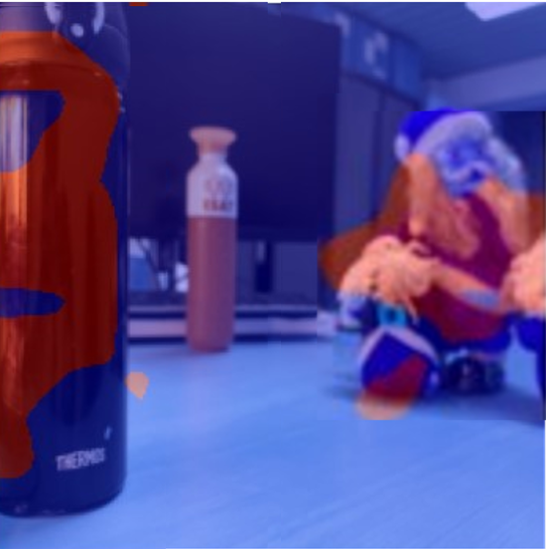}
    }

    \subfigure[Attention map of the left patch in Subfig. 3(c) of the paper.]
    {
          \includegraphics[width=0.45\linewidth]{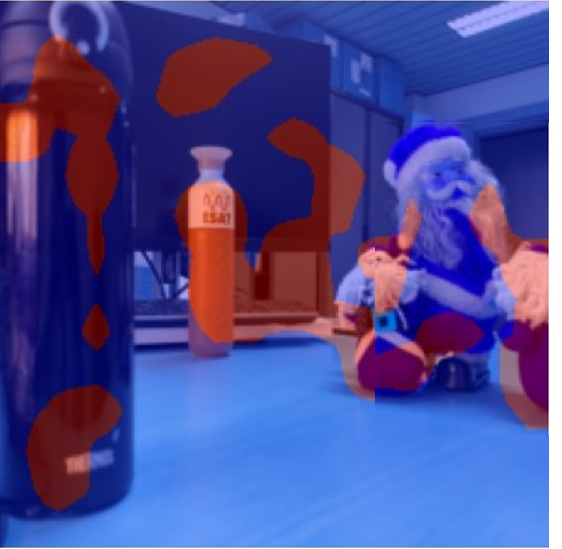}
    }
    \subfigure[Attention map of the right patch in Subfig. 3(c) of the paper.]
    {
          \includegraphics[width=0.45\linewidth]{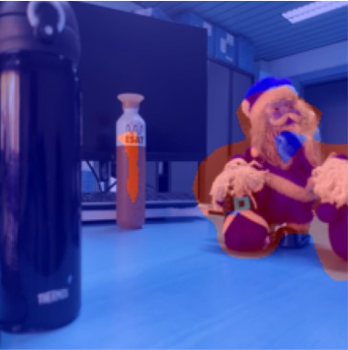}
    }    
    \caption{The full Transformer attention map comparisons of the image input patch in paper main body. 
    \vspace{-0.6em}
}
\label{fig:heatmap-cmps}
\end{figure}

We present the attention heat map of different focus or out-of focus image patched over the whole image in Fig. \ref{fig:heatmap-cmps}. It is obvious that the attention of the Transformer module can attend to more in-focus feature information, \eg, the toy on the right side of Subfig. (d) is with higher attention compared to Subfig. (b) at the same location.  It shows that the patches can attend to the fore- and background regions with related focus and defocus cues of corresponding depth field quite well. It further manifests that the proposed model based on Transformer attention can differentiate the pixel sharpness variance on the image input. Although some attention is put mistakenly like in Subfig. (c), where the attention on the monitor and toy is incorrect, the most of attention is distributed consistently with the input patch appearance. Additionally, the attention of the chosen channel of the Transformer encoder for visualization is not scattered around the whole image, which further discloses that the attention is mainly focused on the similar semantic, sharpness, and appearance information of the input patch.
\subsection{Effects of focal stack size}
\begin{figure}[!thbp]
\centering
\vspace{-0.0em}
\includegraphics[width=0.84\columnwidth]{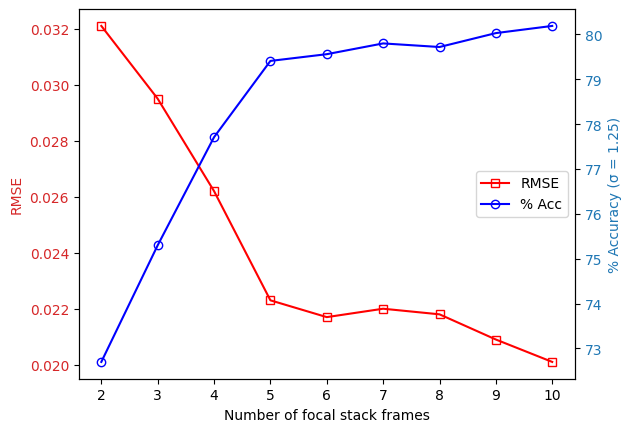}
\caption{Our model performance w.r.t. the frame size of one focal stack sample from DDFF 12-Scene test.}
\vspace{-0.0em}
\label{fig:stack-num}  
\end{figure}

We perform experiments to observe the results of our model on various focal stack sizes. Fig. \ref{fig:stack-num} shows the findings of our experiments on a focal stack from the DDFF 12-Scene validation set, with multiple objects placed at varying depths. In the figure, we plot RMSE and accuracy (in percentage) ($\delta$ = 1.25) w.r.t. the number of focal stack images. We observe that the model accuracy increases with more input images from the focal stack in use, while the RMSE decreases correspondingly. Our model achieves a decent performance around six frames with a notable increase, then followed by a marginal increase after six frames.
To evaluate the impact of token $L_2$ norm threshold, we further tested the RMSE under the various thresholds. The lower threshold smaller than 0.4 like 0.3 can improve the RMSE by a marginal increase at around 0.6\%, yet scaling the model parameter complexity a lot, because more and more tokens are activated, and they are thrown into the LSTM module for processing. The threshold lower than 0.3 can even lead to out of memory issue. Conversely, the large threshold can reduce the LSTM number in use, but also in a sacrifice of the accuracy raging from 0.8\% to 1.2\% as the threshold increased from 0.5 to 0.8.
\section{Experiments}

\subsection{Metrics}
The metrics used in our work for quantitative evaluations, are defined as follows,
\begin{align}
&RMSE: \sqrt{\frac{1}{|M|}\sum_{p\in \mathbf{x}}^{}\lVert {f(\mathbf{x}_{}^{})-\mathbf{D}_{}^{} \rVert}^{2}_{}}, \\
&logRMSE: \sqrt{\frac{1}{|M|}\sum_{p\in \mathbf{x}}^{}\lVert {log f(\mathbf{x}_{}^{})-\mathbf{D}_{}^{} \rVert}^{2}_{}}, \\
&absRel: \frac{1}{|M|}\sum_{p\in \mathbf{x}}^{}\frac{|f(\mathbf{x}_{}^{})-\mathbf{D}_{}|}{\mathbf{D}_{}},\\
&sqrRel  \frac{1}{|M|}\sum_{p\in M}^{}\frac{\lVert f(\mathbf{x}_{}^{})-\mathbf{D}_{}\rVert_{}^{} }{\mathbf{D}_{}},\\
&Bump:  \frac{1}{|M|}\sum_{p\in \mathbf{x}}^{}\min(0.05, \lVert \mathbf{H}_{\Delta}^{}(p) \rVert)\times 100,\\
&Accuracy (\delta): \max \Big(\frac{f(\mathbf{x}_{}^{})}{\mathbf{D}_{}},\frac{\mathbf{D}_{}}{f(\mathbf{x}_{}^{})}\Big)=\delta < threshold, \nonumber \\  & \% \enskip of \enskip \mathbf{D}_{},
\end{align}
where $\Delta = f(\mathbf{x}_{}^{})-\mathbf{D}_{}^{}$ and $\mathbf{H}$ is the Hessian matrix.

\subsection{Network Complexity and Structure Details}
Our network includes the early-stage multi-scale CNN, the Transformer, the LSTM module, and the decoder. We provide a summary of each module's parameter size and its corresponding inference time,
\begin{table}[!thbp]
\centering
\vspace{-1.6em}
\caption{Summary of module size, and inference time for processing one image, for the whole stack, the total time is calculated from a whole stack of images' processing.}
\centering
\begin{tabular}{ c c c }
  \toprule
  Module & Params size &  Inference Time \\ %
  \hline
  Early multi-scale kernel CNNs & 0.487M & 0.001s \\ 
  \hline
  Transformer & 42.065M  & 0.006s \\ %
  \hline
  LSTM & 15.247M & 0.003s\\ %
  \hline   
 CNN Decoder & 16.856M & 0.005s\\ 
  \hline
  \hline
  Total & 74.655M & 0.015\\  
  \bottomrule
\end{tabular}
\label{tab:model_params}
\vspace{-1.8em}
\end{table}
From Tab. \ref{tab:model_params}, we can see the main time consumption is allocated on the Transformer module and decoder, which shows the potential to reduce the model size further can be achieved, \eg, by using MobileViT as an encoder. Our proposed LSTM over the latent feature representation, has a size of only around one-third of Transformer encoder, furthermore, the shallow early multi-scale kernel CNNs is in quite a small size with only 0.487 million parameters and fast processing time around 1ms. The time model size and time complexity summary table further justify our model's compact and efficient design, where the recurrent LSTM module has benefits both in memory size and computational complexity in the design. The main inference time for a single image processing is from the Transformer Encoder, which can be attributed mainly to the attention computation of multiple self-attention heads, and CNN decoder. 

We report the param num in the table below. Although our method contains more parameters (more than half for the Transformer encoder), it increases the performance generally. Our performance gain over the previous competitor is also obvious (with efficient computation as shown in the paper), demonstrating the benefits of using more parameters, \eg, ours gets 0.024 gain on logRMSE using about 2 times param of DFVNet, while DFVNet gets 0.02 gain vs DefocusNet by using about 3 times param of DefocusNet. 

\begin{table}[!thbp]
\vspace{-0.16cm}
\centering
\centering
\caption{Model size and performance comparisons.}
\begin{tabular}{c c c c c}
\hline
&DDFFNet & DefocusNet & DFVNet & Ours\\%
\hline
Param Num & $\approx$40M & $\approx$18M & $\approx$56M & $\mathbf{\approx75M}$ \\ 
\hline
logRMSE & 0.320 & 0.230 & 0.210 & $\mathbf{0.186}$ \\ 
\hline
\end{tabular}
\vspace{-0.8em}
\label{tab:model-complex}
\end{table}
 We present the three main components of our model in Fig. \ref{fig:early_kernel}, Fig. \ref{fig:lstm_transformer}, and Fig. \ref{fig:decoder}, respectively. 
\begin{figure}[!thbp]
\centering
\includegraphics[page=5, trim=0cm 0cm 0cm 0cm, width=0.92\textwidth]{Supplement_figures/Presentation1_supplimentary.pdf}
\caption{The structure details of early-stage multi-scale kernel CNN along with the output size of each layer. The convolution and transpose convolution after concatenation generate a feature map followed by a $1\times1$ convolution along depth channel.}
\label{fig:early_kernel}  
\end{figure}

The early-stage multi-scale CNN in Fig. \ref{fig:early_kernel} takes images from the focal stacks incrementally, \eg a single image $(1, 3, H, W)$ at a time. The $H$, and $W$ denote the image's height and width, respectively. In each figure, the output tensor shape is presented inside the parentheses next to each box. We present each 2D convolution layer with the input feature dimension, output feature dimension, the kernel size and stride inside the parentheses. The early-stage multi-scale CNN generates a feature map of size $(1, 1, H, W)$. The bottom blue boxes in the figure are referred to as "depth channel convolution" in our paper.

\begin{figure}[!thbp]
\centering
\includegraphics[page=6, trim=2cm 17cm 0cm 0cm, width=0.92\textwidth]{Supplement_figures/Presentation1_supplimentary.pdf}
\caption{The LSTM-Transformer block with token size and token dimension.}
\label{fig:lstm_transformer}  
\end{figure}

\begin{figure}[!thbp]
\centering
\includegraphics[page=7, trim=0cm 1cm 0cm 0cm, width=0.92\textwidth]{Supplement_figures/Presentation1_supplimentary.pdf}
\caption{The decoder is composed of repeating composition and fusion modules. The output size of each layer is presented in parenthesis next to the box.}
\label{fig:decoder}  
\end{figure} 

\begin{figure}[!th]
\centering
\includegraphics[trim=0cm 0cm 0cm 0cm, width=1.0\textwidth]{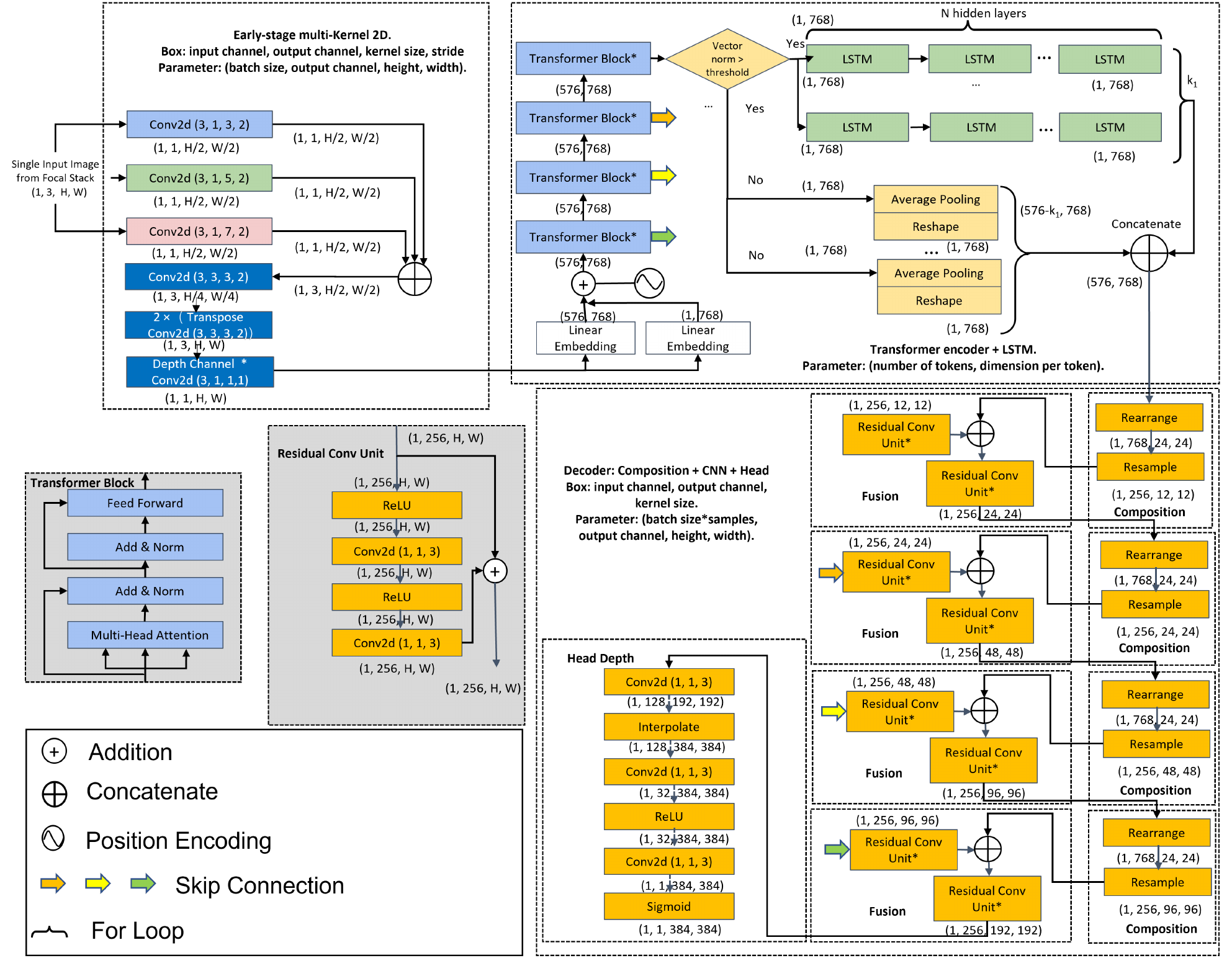}
\caption{The overview structure of our model with notations about the output size of each layer, and the convolution parameters in the parenthesis. The top left is earl-stage multi-scale kernel convolution, the top right is the LSTM-Transformer block,and the bottom right is the decoder, including the repeating compositions and fusions, and a final disparity head.}
\label{fig:overview-structure}  
\end{figure}

The linear embedding layer (Fig. \ref{fig:lstm_transformer}) takes the feature map from early-stage multi-scale kernel convolution, to project into corresponding local patch tokens and a single global token for the whole feature map. The local tokens, the global token, and the position encoding are fused together before being given to the transformer encoder. The output tokens of the last transformer block are grouped based on the vector norm of each token, big norms above a threshold are considered activation tokens, while the rest are regarded as non-activation tokens. The $k_1$ activation tokens are given to the LSTM blocks (Fig. \ref{fig:lstm_transformer}), while we apply average pooling on other $576-k_1$ non-activation tokens. Finally, we add these two groups of tokens to shape the same tokens' arrangement as the input, and then these filtered tokens are delivered to the decoder in Fig. \ref{fig:decoder}. We present the whole network structure in Fig. \ref{fig:overview-structure}.

\subsection{Ablation Study}
\noindent\textbf{Transformer encoder:} Tab. \ref{tab:ddff-encoder-ablation} provides a comparative analysis between Transformer and CNN-based encoders, focusing on the vanilla Transformer, Swin Transformer, and the CNN-based DDFF-Net model used in the whole model structure. 
\begin{table}[!thbp]
\centering
\vspace{-0.0em}
\caption{Metric evaluation of various encoders.}
\label{tab:ddff-encoder-ablation}
\begin{tabular}{c c c c}
\hline
  & RMSE$\downarrow$ & absRel$\downarrow$ & Bump$\downarrow$ \\ 
  \hline
  ViT-base encoder &\bf{2.06e-2} & \bf{0.197} &  \bf{0.29} \\ 
  \hline
  Swin Transformer encoder &2.21e-2 & 0.205 &  0.32  \\ 
  \hline
  CNN encoder & 3.12e-2 & 0.268 & 0.46  \\   
  \hline 
\end{tabular}
\vspace{-1em}
\end{table}

In Tab. \ref{tab:ddff-LSTM-ablation}, we compare our model to its base model without the LSTM module. For the performance without LSTM, the encoder, and decoder are connected directly, and all the depth maps of each image in the stack are averaged to get the metric results of a whole stack. The results indicate the necessity and importance of LSTM in depth estimation from the focal stack problem. It further validates that the modeling stack information separately from the image spatial features can help to improve depth prediction accuracy.

\begin{table}[!thbp]
\centering
\vspace{-0.8em}
\caption{\small{Metric evaluation on different settings for LSTM module on DDFF 12-Scene validation dataset.}}
\begin{tabular}{c c c c}
\hline
  Structure design & RMSE$\downarrow$ & absRel$\downarrow$ & Bump$\downarrow$ \\ 
  \hline    
  Model w/o LSTM module &3.68e-2 & 0.324 & 0.37 \\
  \hline  
  Full model &\bf{1.92e-2} & \bf{0.161} & \bf{0.19}  \\
  \hline
\end{tabular}
\label{tab:ddff-LSTM-ablation}
\vspace{-2.8em}
\end{table}

\subsection{Visual results}
\begin{figure}[!thbp]
\centering
\begin{tabularx}{\textwidth}{c c c c c c c}
    \hspace{1.2cm} Input & \hspace{1.0cm} GT & \hspace{0.8cm} DDFFNet & \hspace{0.2cm} DefocusNet & \hspace{0.4cm}AiFNet & \hspace{0.8cm}DFVNet & \hspace{0.8cm} Ours  \\
\end{tabularx}
\includegraphics[page=1, trim=0cm 2cm 0cm 0cm, width=1.0\textwidth]{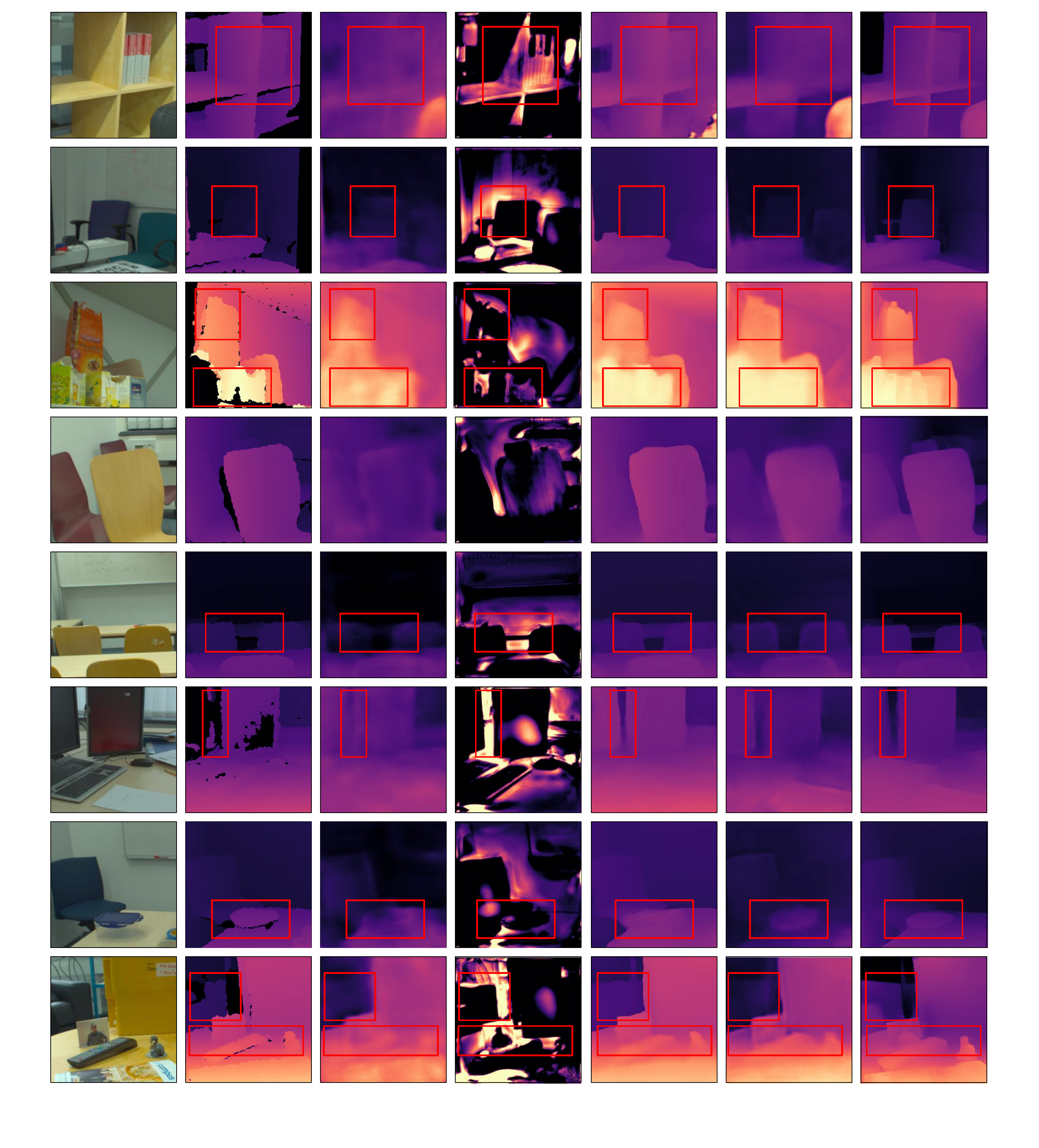}
\caption{Qualitative evaluation of our model on DDFF 12-Scene dataset validation set.}
\label{fig:ddff-viz}  
\end{figure}

In this section, we present the additional visual comparison results of our model on DDFF 12-Scene, Mobile Depth, and LightField 4D datasets in Fig. \ref{fig:ddff-viz}, Fig. \ref{fig:fod-viz}, Fig. \ref{fig:mobiledepth-viz}, Fig. \ref{fig:lightfield-viz}, respectively. All comparison results include ground truths except the Mobile Depth dataset, captured by cellphones. 

\noindent\textbf{Visual Results on DDFF 12-Scene Dataset.}
Fig. \ref{fig:ddff-viz} shows examples of depth prediction of our model on an additional set of samples from the DDFF 12-Scene validation set. We show that our model preserves a lot of local details on disparity map, \eg, books on the shelf (first row), the chair (second row), and the black-colored couch in the top-left corner (last row). Our model predicts an accurate depth with sharper and smoother boundaries compared to prior works.

\noindent\textbf{Visual Results on FOD500 Dataset.}
We report our qualitative results on the last 100 images of the FOD-500 dataset in Fig. \ref{fig:fod-viz} to compare them with prior works. We observe decent disparity prediction results with smoother boundaries and preserved local structure details, \eg, the center hole (last row) and the small hole at the top center (sixth row). The DDFFNet and DefocusNet are fine-tuned on FOD-500 with their weights pre-trained on DDFF 12-Scene dataset. We achieve comparable performance to DFVNet and AiFNet, yet better than other prior works.

\noindent\textbf{Visual Results on Mobile Depth Dataset.} Fig. \ref{fig:mobiledepth-viz} depicts the additional visual results of our model on Mobile Depth dataset for cross-dataset evaluation on \emph{varying focal stack length}. The figure shows that our model consistently predicts accurate disparity from an arbitrary number of images in focal stacks, although it has never been trained on this dataset. In comparison to prior works, our model achieves better estimations that predict sharp and crisp boundaries, \eg, the metal tap (second row), the plant (third row) and the ball next to the window (last row). Also the mirror effects pose a big challenge to our model, as visualized in the last row at the window background. 

\begin{figure}[!thbp]
\centering
\includegraphics[width=0.26\linewidth]{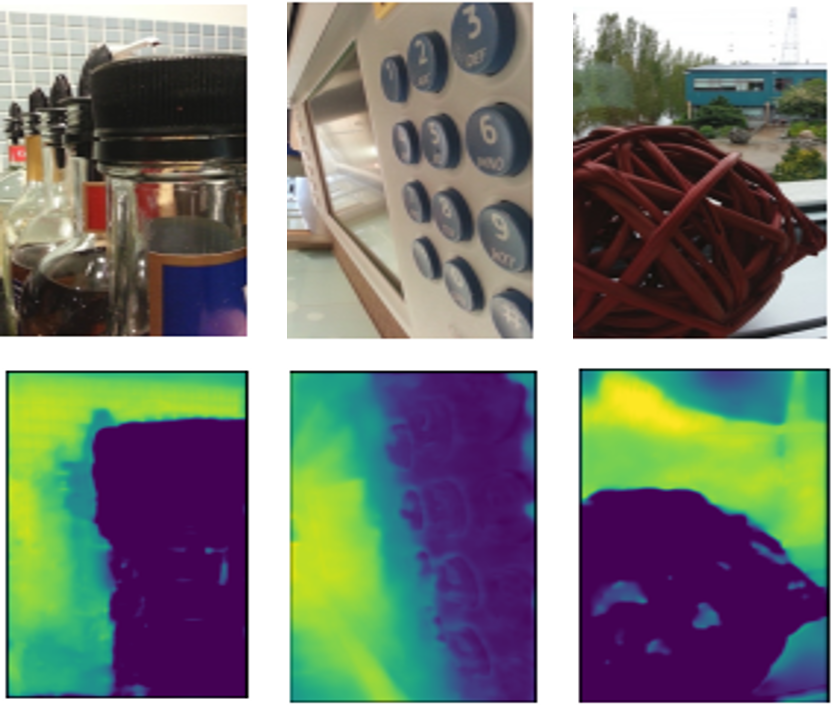}
\centering
\caption{Qualitative evaluation of DEReD \cite{si2023fully} on Mobile Depth, with the original visual results from the paper.}
\label{fig:mobile-dered}  
\end{figure}

For the latest DEReD \cite{si2023fully} model which estimates the all-in-focus image and depth map jointly. And as claimed in the paper, the results seem to be quite plausible, so we also tried to implement its publicly available code, however we found the released code is not complete with some self-defined modules missing, and the configuration of the model is quite difficult to understand and hard to provide all the required input with a lot of hard-coded parameters, so we cannot reproduce or re-train the model on our comparable datasets. Furthermore the metric results used in the paper are evaluated on DefocusNet dataset and post-processed NYUv2 dataset by using its proposed rendering defocus image tool, which is different from the normal focal stack benchmark settings as used in our paper, so we only provide the original qualitative comparison results on Mobile Depth from the paper in Fig. \ref{fig:mobile-dered} as a reference.

\noindent\textbf{Visual Results on LightField4D Dataset.}
We present an additional set of visual results of our model on LightField4D dataset in Fig. \ref{fig:lightfield-viz}. More specifically, we employ the "additional set" from the dataset for cross-dataset evaluation. Our model estimates better depth on this unseen dataset, that preserves the local details in many complex structures, \eg, the tower (first row), board game (second row), and the tomb (fifth row). Our model achieves a deficit performance in shadow, and textureless scenes, such as the pillow (last row) and "Antinous" (sixth row).

\begin{figure}[!thbp]
\centering
\begin{tabularx}{\textwidth}{c c c c c c c}
    \hspace{1.2cm} Input & \hspace{1.0cm} GT & \hspace{0.8cm} DDFFNet & \hspace{0.2cm} DefocusNet & \hspace{0.4cm}AiFNet & \hspace{0.8cm}DFVNet & \hspace{0.8cm} Ours  \\
\end{tabularx}
\includegraphics[page=2, trim=0cm 0cm 0cm 0cm, width=1.0\textwidth]{Supplement_figures/Presentation1_supplimentary.pdf}
\vspace{-2.8em}
\caption{Qualitative evaluation of our model on FOD500 dataset. Only the last 100 focal stacks are used for testing. DFVNet uses first 400 focal stacks for training and other models use the same split for fine-tuning.}
\label{fig:fod-viz}  
\end{figure} 

\begin{figure}[!thbp]
\centering
\begin{tabularx}{\textwidth}{c c c c c c c}
    \hspace{2.4cm} Input & \hspace{0.4cm} DDFFNet & \hspace{-0.00cm} DefocusNet & \hspace{0.6cm}AiFNet & \hspace{0.6cm}DFVNet & \hspace{0.6cm} Ours  \\
\end{tabularx}
\includegraphics[page=3, trim=0cm 8cm 0cm 0cm, width=0.95\textwidth]{Supplement_figures/Presentation1_supplimentary.pdf}
\vspace{-1.8em}
\centering
\caption{Qualitative evaluation of our model on Mobile Depth dataset. This dataset contains focal stacks of varying lengths and image sizes, we cropped raw image inputs from the top left.}
\label{fig:mobiledepth-viz}  
\end{figure} 

\begin{figure}[!thbp]
\centering
\begin{tabularx}{\textwidth}{c c c c c c c}
    \hspace{1.2cm} Input & \hspace{1.0cm} GT & \hspace{0.8cm} DDFFNet & \hspace{0.2cm} DefocusNet & \hspace{0.4cm}AiFNet & \hspace{0.8cm}DFVNet & \hspace{0.8cm} Ours  \\
\end{tabularx}
\includegraphics[page=4, trim=0cm 0cm 0cm 0cm, width=1.0\textwidth]{Supplement_figures/Presentation1_supplimentary.pdf}
\caption{Qualitative evaluation of our model on an additional set of LightField4D dataset.}
\label{fig:lightfield-viz}  
\end{figure} 

